\documentclass{article} 
\usepackage{iclr2026_conference,times}

\usepackage{amsmath,amsfonts,bm}

\def\eqref#1{equation~\ref{#1}}

\def\1{\bm{1}}

\DeclareMathAlphabet{\mathsfit}{\encodingdefault}{\sfdefault}{m}{sl}
\SetMathAlphabet{\mathsfit}{bold}{\encodingdefault}{\sfdefault}{bx}{n}

\usepackage{hyperref}       
\usepackage{url}            
\usepackage{booktabs}       
\usepackage{multirow}
\usepackage{amsfonts}       
\usepackage{nicefrac}       
\usepackage{microtype}      
\usepackage[table]{xcolor}         
\usepackage{amsmath}
\usepackage{subcaption}
\usepackage{graphicx}
\usepackage{xspace}
\usepackage{wrapfig}

\definecolor{darkred}{RGB}{139,0,0}
\definecolor{darkblue}{RGB}{0,0,139}
\definecolor{lightblue}{RGB}{220,230,250}

\newcommand{\rebuttal}[1]{\textcolor{black}{#1}}

\newcommand{\full}{Item-ID + Natural-language Mixture-of-Experts Language Model\xspace}
\newcommand{\abbr}{IDIOMoE\xspace}

\title{Catalog-Native LLM: Speaking Item-ID dialect with Less Entanglement for Recommendation}

\author{
Reza Shirkavand\thanks{Work done during internship at Roblox}~$^{1}$, Xiaokai Wei$^{2}$, Chen Wang$^{2}$, Zheng Hui$^{*3}$, Heng Huang$^{1}$ , Michelle Gong$^{2}$ \\
$^{1}$University of Maryland - College Park \footnotesize \texttt{\{rezashkv,heng\}@cs.umd.edu},\\
$^{2}$Roblox \footnotesize \texttt{\{xwei,cwang,mgong\}@roblox.com}\\
$^{3}$University of Cambridge \footnotesize \texttt{zh2483@columbia.edu}\\
}

\iclrfinalcopy 
\begin{document}

\maketitle

\begin{abstract}
While collaborative filtering delivers predictive accuracy and efficiency, and Large Language Models (LLMs) enable expressive and generalizable reasoning, modern recommendation systems must bring these strengths together. Growing user expectations, such as natural-language queries and transparent explanations, further highlight the need for a unified approach. However, doing so is nontrivial. Collaborative signals are often token-efficient but semantically opaque, while LLMs are semantically rich but struggle to model implicit user preferences when trained only on textual inputs.
This paper introduces \full(\abbr), which treats item interaction histories as a native dialect within the language space, enabling collaborative signals to be understood in the same way as natural language. By splitting the Feed Forward Network of each block of a pretrained LLM into a separate text expert and an item expert with token-type gating, our method avoids destructive interference between text and catalog modalities. 
\abbr demonstrates strong recommendation performance across both public and proprietary datasets, while preserving the text understanding of the pretrained model.
\end{abstract}

\section{Introduction}\label{sec:introduction}

Recommendation systems shape what people read, watch, buy, learn, and play. As AI shifts from static predictors to reasoning agents capable of following instructions, recommendation is also evolving from ranking fixed lists to assisting users in exploring, planning, and deciding. This trend is visible in practice: Amazon’s Rufus provides LLM-powered conversational shopping~\citep{amazon_rufus2024}; Meta’s Llama-3 assistant is embedded in WhatsApp, Instagram, and Facebook for task planning~\citep{meta_llama3_2024}; and Netflix is adopting foundation-model approaches for personalization and LLM-based conversational retrieval~\citep{netflix_fm_2025,zhu2025collaborativeretrievallargelanguagenetflix}. These examples motivate bringing LLM knowledge and instruction-following into recommenders while preserving the collaborative patterns that make them accurate at scale.

Conventional recommenders like collaborative filtering (CF)\citep{koren2009matrix}, content-based (CB)\citep{Lops2011content}, and sequential models~\citep{kang2018sasrec,sun2019bert4rec,zhai2024hstu} perform well within their scope when data are abundant, but they depend heavily on the quality of logs and item attributes. They remain vulnerable to popularity bias~\citep{abdollahpouri2019managing-pop-bias}, struggle to integrate heterogeneous signals (text, behavior, and context), and cannot support natural language queries.

Pre-trained LLMs offer complementary strengths: they bring broad world knowledge, can follow natural-language instructions, and can reason about multi-objective trade-offs. Yet a fundamental gap remains. LLM pretraining centers on semantic understanding, whereas recommendation requires modeling collaborative preference patterns. The key challenge is leveraging LLMs for preference understanding without disrupting their semantic competence.

Recent work has tried to bridge this gap by extending LLM vocabularies with item IDs~\citep{cao2024gdm-aligning,zhu2024ccllm4rec,jiang2025urm, zhang2025cove}, enabling direct ID-level generation. While effective in principle, such naive integration often causes knowledge interference: collaborative signals entangle with linguistic semantics, leading to degraded performance on both sides. As we'll show, this interference does not vanish by simply scaling up parameters (e.g. adding more parameters naively) and thus calls for more principled architectural solutions.

\begin{figure}[t]
  \centering
  \begin{subfigure}[t]{0.243\linewidth}
    \includegraphics[width=\linewidth]{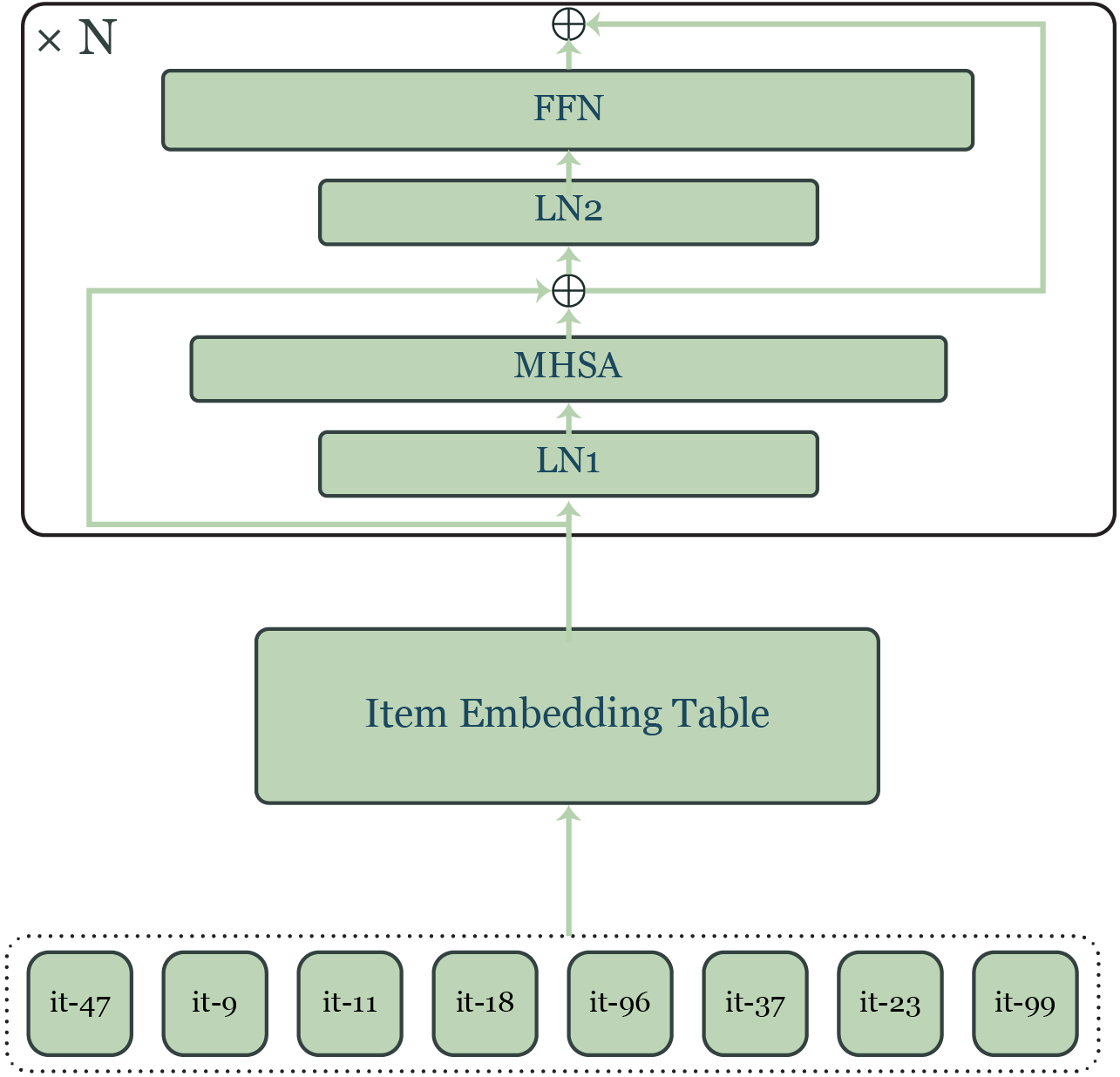}
    \caption{}
    \label{fig:teaser-a}
  \end{subfigure}\hfill
  \begin{subfigure}[t]{0.24\linewidth}
    \includegraphics[width=\linewidth]{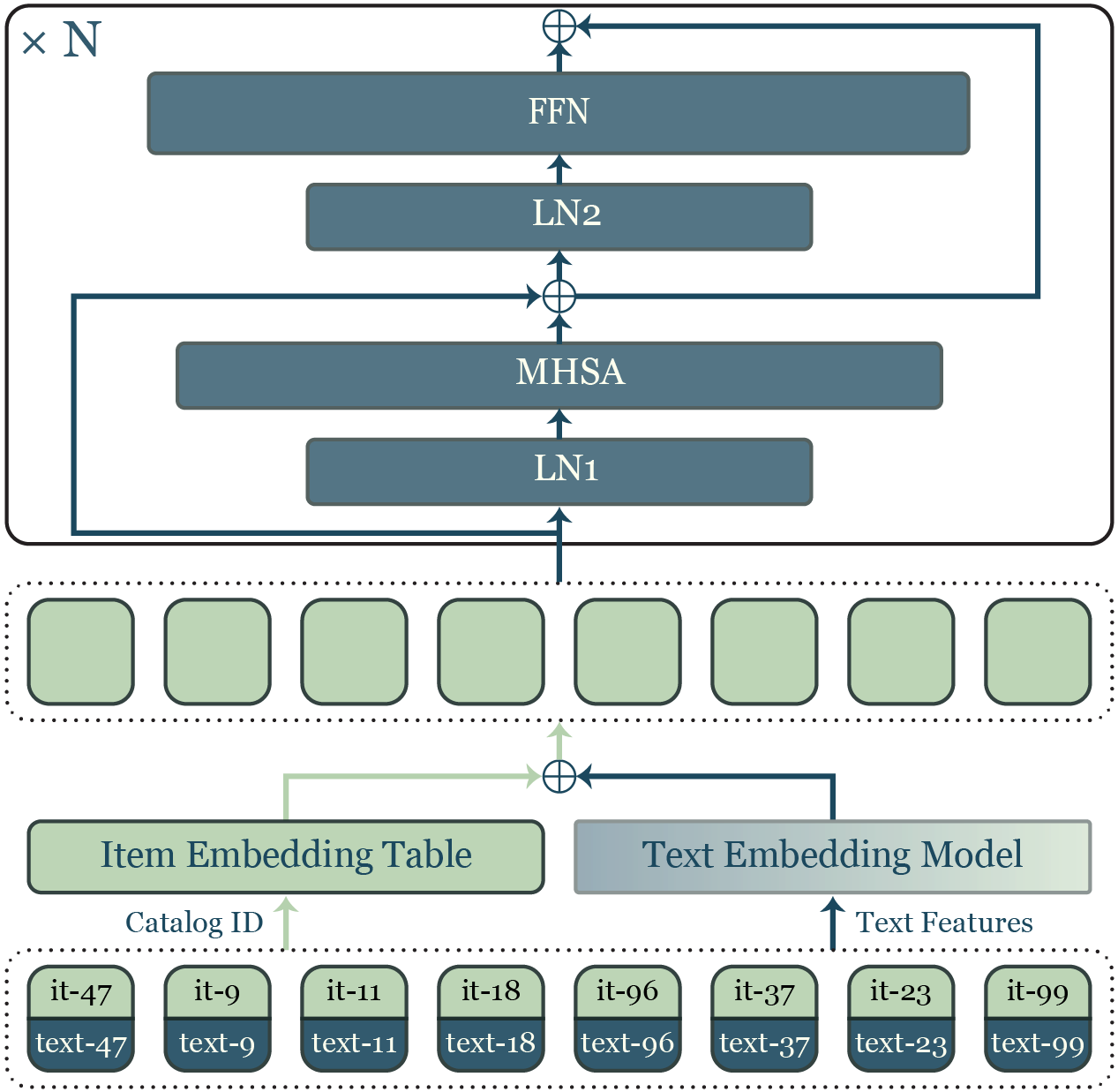}
    \caption{}
    \label{fig:teaser-b}
  \end{subfigure}\hfill
  \begin{subfigure}[t]{0.24\linewidth}
    \includegraphics[width=\linewidth]{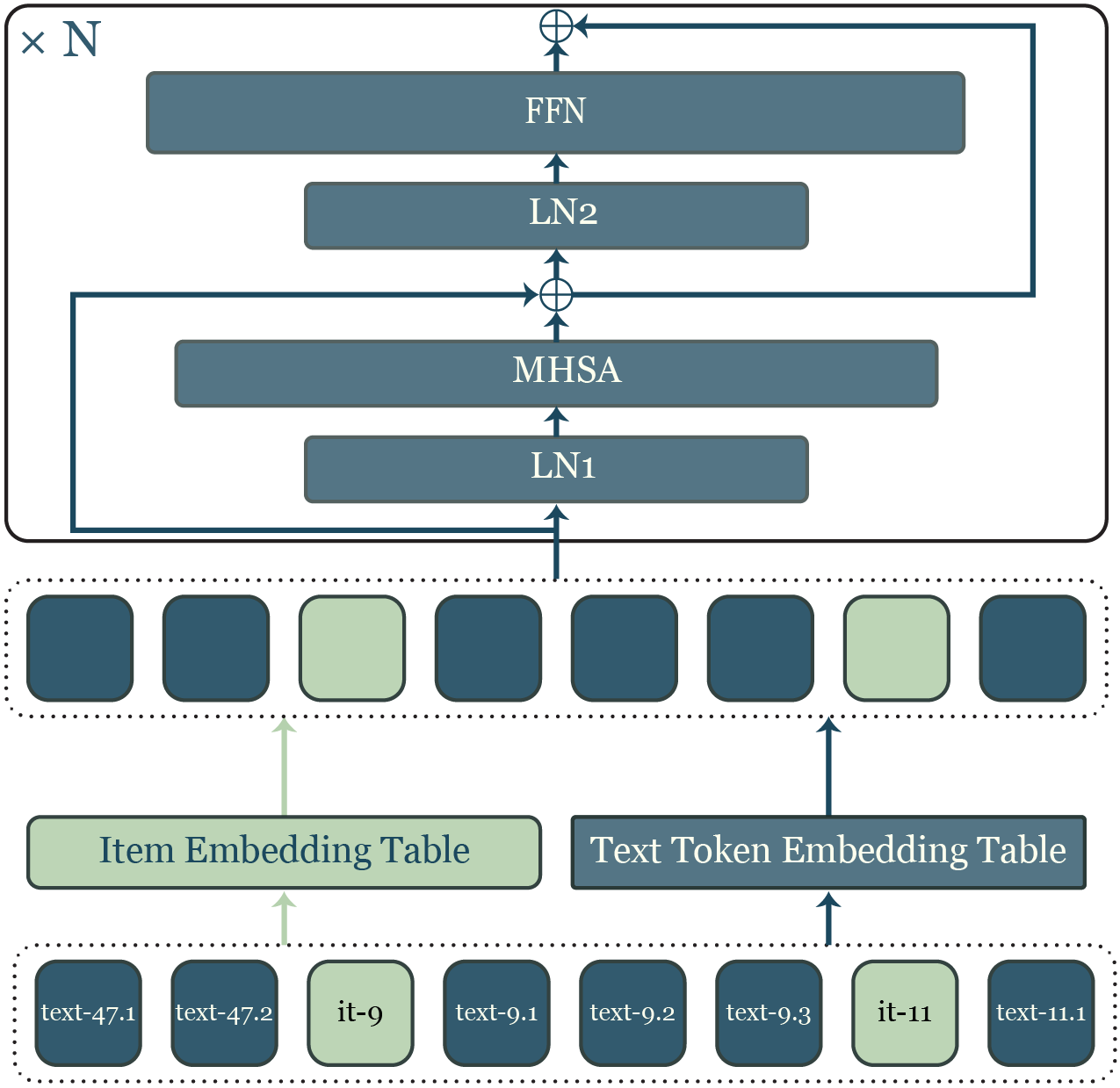}
    \caption{}
    \label{fig:teaser-c}
  \end{subfigure}\hfill
  \begin{subfigure}[t]{0.24\linewidth}
    \includegraphics[width=\linewidth]{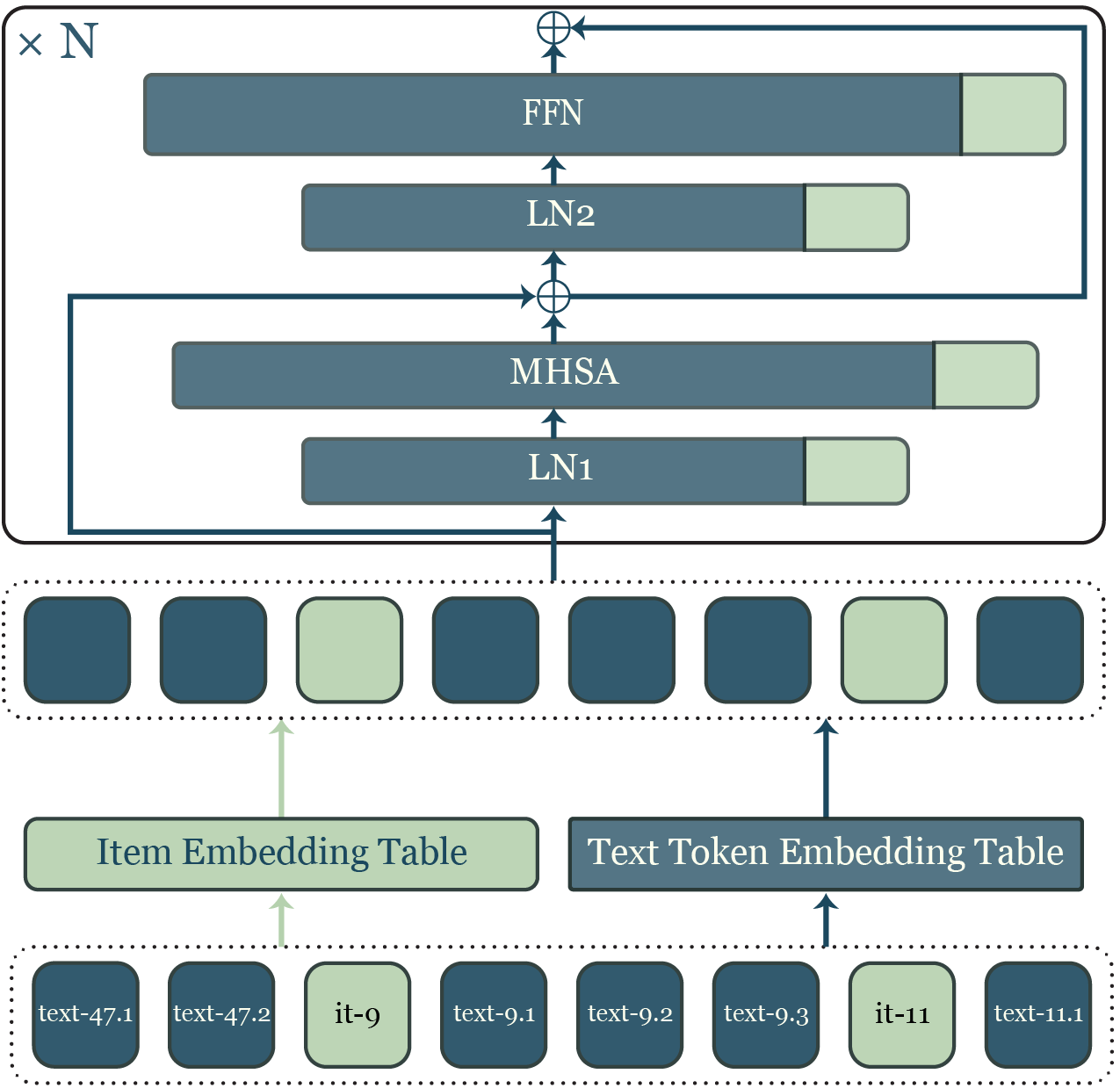}
    \caption{}
    \label{fig:teaser-d}
  \end{subfigure}
  \caption{Four designs for recommendation with Transformers/LLMs. (a) ID-only Transformer: trained from scratch on item-ID sequences, with no pretrained LLM involved. (b) Text-derived bias: a pretrained LLM on IDs, with an external text encoder providing side features that bias item scores. (c) Explicit text tokens: a pretrained LLM that directly consumes both item-ID tokens and (possibly) text tokens in the same sequence. (d) Explicit text tokens + extra capacity: like (c), but adds item-specific parameters to better handle IDs. \abbr{} is a special case of (d).}
  \label{fig:teaser}
\end{figure}

Inspired by mixture-of-experts (MoE)~\citep{shazeer2017outrageously,lepikhin2020gshard,fedus2022switch}, we view ItemID modeling as a dialect distinct from natural language. But unlike standard MoE, which routes tokens indiscriminately, we design a targeted \textit{\full(\abbr)} that assigns a dedicated collaborative expert for IDs alongside a preserved text expert for language. A token-type gate orchestrates their interaction, mitigating interference while retaining pretraining knowledge. 
Evaluations on both public benchmarks and a real-world industrial dataset from a leading online platform with hundreds of millions of users show that \abbr consistently outperforms text-only adapters and item-only baselines. Our main contributions are:

\paragraph{Disentangled MoE architecture for recommendation.}  
We propose a Mixture-of-Experts design that treats Item-IDs as a native dialect. To the best of our knowledge, this is the first attempt at separating collaborative filtering from semantic processing, with a router that activates text experts only when useful.  

\paragraph{Robust performance on real-world scale.}  
Our method achieves compelling results on public datasets and on our large proprietary dataset with more hundreds of millions of users, while maintaining the natural language understanding of a pre-trained LLM.

\paragraph{Extensive ablations isolating the source of gains.}  
We study model capacity and matched-capacity non-MoE baselines showing that improvements arise from expert specialization and routing, not just added parameters. 

\paragraph{Analysis of expert specialization.}  
Through a key-value memory lens of FFN neurons, we show that MoE separation yields clearer item-text affinity, higher category purity, and more clustered neurons than a non-MoE baseline, providing evidence that expert disentanglement leads to more interpretable and modular representations.

\section{Related Work}\label{sec:related-work}
\subsection{Conventional Recommendation Methods}

Traditional recommendation models fall into collaborative filtering (CF), content-based (CB), and sequential paradigms. CF learns from user–item interactions to model latent preferences~\citep{koren2009matrix}, while CB leverages item attributes to improve personalization and mitigate cold-start issues~\citep{Lops2011content}. Sequential models further capture temporal dynamics, using models such as RNNs~\citep{hidasi2015gru4rec}, SASRec~\citep{kang2018sasrec}, and BERT4Rec~\citep{sun2019bert4rec}. Though these models achieve strong performance under sufficient data, they operate on opaque ID sequences and require hand-crafted features or specialized architectures to incorporate diverse signals like language or intent. They also struggle with long-tail exposure~\citep{abdollahpouri2019managing-pop-bias}.

\subsection{Generative Recommendation }
Some works treat recommendation as sequence generation, unifying retrieval and ranking under a generative objective~\citep{yang2025gr-llm}. This includes large-scale decoder models such as HSTU~\citep{zhai2024hstu}, which scales to trillions of parameters, and OneRec~\citep{deng2025onerec}, which uses a sparse MoE encoder–decoder architecture for scalable training. \rebuttal{These approaches improve novelty, fluency, and explainability, but are resource-intensive and require careful objective and data design to fully exploit collaborative interaction signals.} They also do not support conversational recommendation.

\subsubsection{LLM-Based Recommendation and Semantic–ID Alignment}
Large language models (LLMs) offer world knowledge and instruction-following capabilities that are appealing for building explainable recommenders. Recent frameworks such as P5~\citep{geng2022RLP} reframe recommendation tasks as text-to-text generation, supporting few-shot generalization. Prompt-based methods~\citep{hou2024llm-zeroshot-ranker} further explore LLMs as zero-shot rankers. However, these methods require verbose text inputs and often discard raw user–item interaction data, missing collaborative patterns entirely. To bridge this semantic collaborative gap, prior work fine tunes on interactions~\citep{cao2024gdm-aligning}, aligns with rewards~\citep{lu2024aligning-llm-rl}, or unifies modalities in shared token spaces~\citep{zhai2025mm-quant-gen}. A complementary direction embeds item IDs as tokens (e.g., CoVE~\citep{zhang2025cove}, CLLM4Rec~\citep{zhu2024ccllm4rec}, URM~\citep{jiang2025urm}), enabling token efficient generation and retrieval. However, designs like URM that drop explicit text tokens, hinder conversational recommendation and instruction handling. And when ID tokens and text tokens share parameters, interference emerges: language and collaborative signals entangle, degrading both.

\subsection{Multimodal MoE LLMs}
Recent work integrates MoE into multimodal LLMs (MLLMs) and LVLMs \cite{bao2022vlmounifiedvlp,shen2023scalingvlmmoe,diao2025eve2,deng2025bagel}. MoE-LLaVA~\citep{lin2024moellava} adds a sparse MoE backbone to LLaVA~\citep{liu2023llava}, converting feed-forward blocks into experts to match or exceed larger dense variants while activating fewer parameters and reducing visual hallucinations. Uni-MoE~\citep{li2025unimoe} scales unified multimodal LLMs across many modalities and tasks with MoE layers. MoME~\citep{mome} further mitigates task interference by factorizing the model into a Mixture of Vision Experts (MoVE) and a Mixture of Language Experts (MoLE), with MoVE aggregating multi-encoder vision features via an instruction-conditioned router and MoLE using sparsely gated adapter experts.

\subsection{Motivation and Positioning}

\begin{figure}[t]
    \centering
    \includegraphics[width=\linewidth]{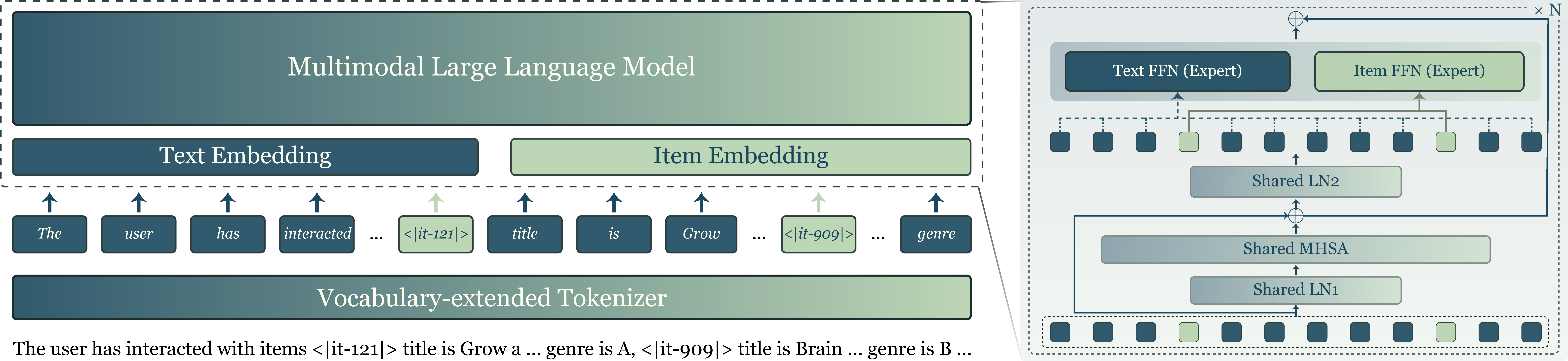}
    \caption{Overview of our proposed \abbr. We extend the LLM tokenizer with new \texttt{"item-id"} tokens and introduce a dedicated item embedding layer. The Normalization and Attention layers are shared across all token types, while tokens are routed to distinct FFN layers depending on their type.}
    \label{fig:method}
\end{figure}

While prior work has shown the potential of combining semantic understanding with collaborative signals, existing methods lack clear mechanisms to separate and preserve these distinct forms of knowledge. Text can be incorporated via (a) \emph{text-as-features} (pre-encoded embeddings/biases attached to IDs); or (b) \emph{explicit text tokens} (Figure~\ref{fig:teaser}). We choose the latter to preserve conversational capabilities of the LLM.
 In this setting, interference between language understanding and ID-level preference modeling remains an underexplored bottleneck. Simply mixing tokens or scaling capacity does not solve it.

We address this challenge by introducing a \textit{\full(\abbr)} that treats item interactions as a native dialect. 
\abbr dedicates separate pathways to item and text processing in each block, with a lightweight token-type gate that reduces interference while retaining language understanding. This design enables efficient ID-level modeling and better alignment with both semantic and collaborative objectives.

\section{Method}\label{sec:method}

\subsection{Preliminary}\label{sec:method_sub:prelim}

We study how incorporating item textual attributes affects performance given a user's interaction history. We start from the pretrained \texttt{Qwen/Qwen2.5-0.5B}~\citep{qwen2025qwen25technicalreport}, extend its vocabulary with item-ID tokens, and compare three variants that differ only in input format and the source of item embeddings. In all variants, instruction text tokens are embedded with the LLM’s native token embedding matrix.

\begin{enumerate}
    \item \textbf{ID-only (learned ID embeddings).} Input: \textit{``The user has interacted with \texttt{<|item-53|> <|item-11|>} ...''}. Each item token is embedded via a learned item embedding table.

    \item \textbf{ID-only + text-derived bias.} Following~\citet{jiang2025urm} this variant has same input as (a). However, each item token embedding is the sum of (i) a learned ID vector and (ii) a text-derived vector computed from the item’s title and category using a general-purpose sentence-embedding model.

    \item \textbf{ID + explicit attributes.} Input interleaves IDs with attributes: \textit{``The user has interacted with \texttt{<|item-53|>} title: \texttt{X}, category: \texttt{Y}; \texttt{<|item-11|>} ...''}. Item-ID tokens use the learned item embedding table; Text tokens are embedded by the LLM’s token embeddings.
\end{enumerate}

\begin{table}
\centering

\begin{minipage}{0.45\linewidth}
\centering
\vspace{-15pt}
\captionof{table}{Improvements over the ID-only baseline when adding text features.}
\label{tab:text-feat-ablation}
\resizebox{\linewidth}{!}{%
\begin{tabular}{lcccc}
\toprule
\multirow{2}{*}{\textbf{Variant}} & \multicolumn{2}{c}{\textbf{Arts} $\Delta(\%)$} & \multicolumn{2}{c}{\textbf{Industrial} $\Delta(\%)$} \\
\cmidrule(lr){2-3}\cmidrule(lr){4-5}
 & HR@10 & NDCG@10 & HR@10 & NDCG@10 \\
\midrule
ID-only (baseline)                 & \multicolumn{4}{c}{---} \\
ID-only + text-derived bias        & +42.8\% & +26.4\% & +18.1\% & +13.9\% \\
ID + explicit attributes           & +24.6\% & +17.6\% & +11.4\% & +6.8\% \\ \midrule
\textbf{\abbr}                              & \textbf{+44.1}\% & \textbf{+28.1}\% & \textbf{+22.7}\% & \textbf{+14.2}\% \\ 
\bottomrule
\end{tabular}
}
\end{minipage}
\hfill
\begin{minipage}{0.54\linewidth}
\centering
\captionof{figure}{Language understanding retention.}
\includegraphics[width=\linewidth]{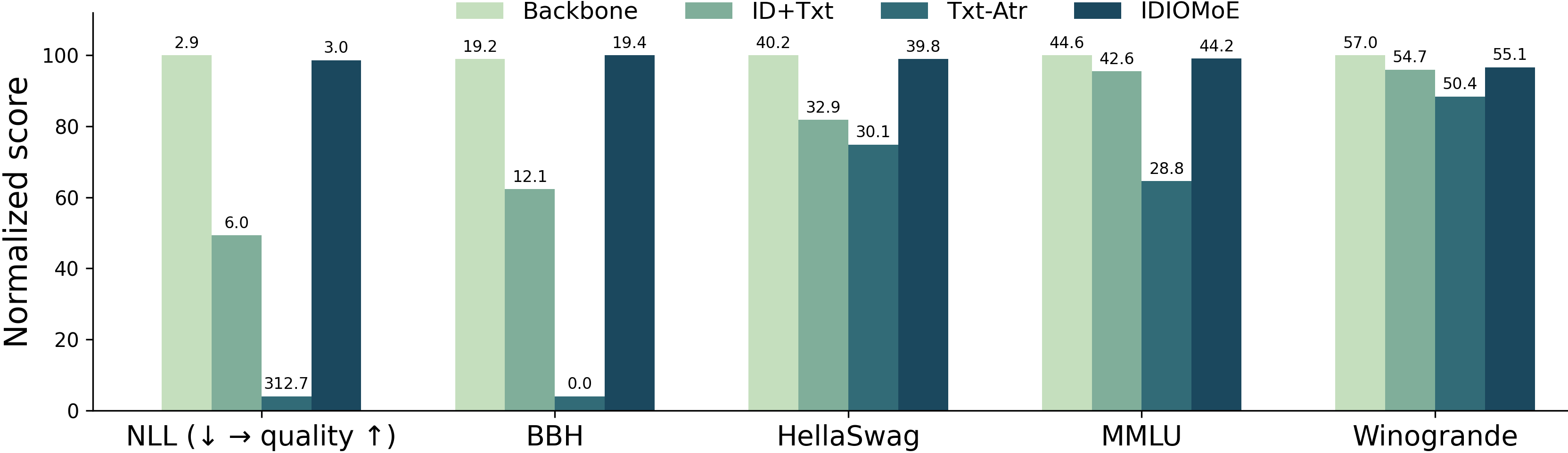}
\label{subfig:nll-upright}
\label{fig:nll-figure}
\end{minipage}
\end{table}

We evaluate the above on two datasets: Amazon-Arts~\citep{ni2019amzn} and our industrial dataset. The results are presented in Table~\ref{tab:text-feat-ablation}.
In both datasets adding item textual attributes improves performance. The text-derived bias approach performs better as it is easier for the model to handle as it adds some semantic signal without making the sequence longer or more complex. In contrast, giving the model full attribute text makes the input longer and harder to learn from. But there is a key reason to still include explicit text: it enables capabilities that the bias method can’t. Conversing with users and
generating user-friendly explanations all rely on having real text.

To evaluate whether the variants preserve the pretrained model’s linguistic ability, we measure negative log-likelihood (NLL) on 5,000 samples from the \texttt{wikitext} validation set~\citep{merity2016wikitext} and further assess performance on four benchmarks: BBH~\citep{suzgun2022bbh}, HellaSwag~\citep{zellers2019hellaswag}, MMLU~\citep{hendryckstest2021mmlu}, and WinoGrande~\citep{sakaguchi2019winogrande}. As shown in Figure~\ref{fig:nll-figure}, ID+Text achieves substantially lower NLL and significantly higher benchmark results compared to the text-derived bias variant. While the bias method provides strong recommendation accuracy, it does so at the cost of language degradation, reflected in much poorer performance on language understanding tasks. This points to the need for a better approach; one that preserves the advantages of explicit text for conversational recommendation while still achieving strong performance on standard recommendation tasks.

In this paper, we propose to divide responsibilities rather than forcing a single model to handle everything. One expert is dedicated to IDs and collaborative filtering, while another is responsible for text. This design allows us to retain the benefits of explicit text when needed, without sacrificing efficiency or accuracy when it is not. We show \abbr preserves the language understanding of the model, while delivering the best recommendation performance~(Table~\ref{tab:text-feat-ablation} and Figure~\ref{fig:nll-figure}), confirming that separating experts by token type reduces semantic–collaborative interference.

\subsection{\abbr}\label{sec:method_sub:ours}
We present the \emph{\full} (\abbr), a pretrained decoder-only LLM augmented with item-specialized experts and native item tokens. \abbr keeps the language skills of the base model intact while learning collaborative patterns directly from user-item sequences.
We start from a pretrained causal transformer and replace each feed-forward network (FFN) with a two-expert module:
\begin{itemize}
  \item \textbf{Text Expert}: the original FFN from the pretrained LLM, preserved as-is.
  \item \textbf{Item Expert}: a new FFN similar to  the text expert, optionally shrunk (e.g., $\times\!\frac{1}{2}, \times\!\frac{1}{4}$) to add capacity efficiently.
\end{itemize}

Routing is handled by a \textbf{static token-type gate}:
\rebuttal{We use a simple static routing scheme: only item-ID tokens \texttt{<|it-.|>} are routed to the item expert, and all other tokens (titles, attributes, etc.) are routed to the text expert. All tokens share the same self-attention layers at every depth, so IDs and text always attend to each other, and the MoE split only affects the FFN sublayers, i.e., where ID- vs. text-specific information is stored. This design lets the model jointly reason over blended textual attributes and item IDs while allocating separate capacity for catalog structure.}
 Moreover, since only one expert is active per token, so compute stays comparable to the base model (\rebuttal{See Appendix~\ref{app:efficiency-results} for a discussion of efficiency results}). Figure~\ref{fig:method} provides an overview of our framework.

\subsubsection{Native item tokens and hybrid head.}
We augment the tokenizer with special item tokens \texttt{<|it-\emph{id}|>} and attach a hybrid embedding layer that combines the frozen text embeddings with a trainable item embedding table. The output head reuses the same hybrid parameterization so the model can generate item IDs directly.

\subsection{FFN Key-Value Memory Analysis}\label{sec:ffn-kv-analysis}

\subsubsection{Setup}  
Following~\citet{geva2022key-val-memory-analysis}, we view each feed-forward network (FFN) in a transformer block as a key-value memory, where hidden states act as queries and FFN neurons contribute value vectors. Our goal is to probe whether Mixture-of-Experts (MoE) separation encourages the \emph{item expert} to encode item semantics distinct from the \emph{text expert}, and how this differs from a non-MoE baseline.

For a transformer layer $\ell \in \{1,\dots,L\}$, let the FFN consist of two linear projections with activation in between. We denote the second projection as  
$W^{(\ell)}_{\text{out}} \in \mathbb{R}^{I \times d}$
where $I$ is the FFN hidden dimension and $d$ is the model dimension. Each row $w^{(\ell)}_j \in \mathbb{R}^d$ of $W^{(\ell)}_{\text{out}}$ is treated as a \emph{value vector} associated with neuron $j$ in layer $\ell$. To study how these rows align with model embeddings, we construct two sets of reference vectors:

\begin{itemize}
  \item \textbf{Item embeddings:} $E_{\text{items}} \in \mathbb{R}^{N_{\text{items}} \times d}$, taken from the learned item embedding table used for ID tokens.
  \item \textbf{Text token embeddings:} $E_{\text{text}} \in \mathbb{R}^{V_{\text{text}} \times d}$, taken from the backbone’s input embedding matrix for standard vocabulary tokens (excluding items).
\end{itemize}

Given a value vector $w \in \mathbb{R}^d$, we compute cosine similarities to both sets:  
\begin{equation}
    s_{\text{items}}(w) = E_{\text{items}} w^\top, 
\quad
s_{\text{text}}(w) = E_{\text{text}} w^\top,
\end{equation}
assuming all vectors are $\ell_2$-normalized. We then retrieve the top-$k$ most similar item IDs and text tokens for analysis.

\subsubsection{Metrics}  
We define three metrics to quantify the specialization of each neuron’s value vector $w$:  

\begin{equation}
    \textbf{Affinity:}\quad
a(w) = \operatorname{median}\!\big(s_{\text{items}}^{\text{top-}k}(w)\big)\;-\;\operatorname{median}\!\big(s_{\text{text}}^{\text{top-}k}(w)\big), \label{eq:affinity}
\end{equation}

\begin{equation}
\textbf{Purity:}\quad
p(w) = \max_{c \in \mathcal{C}} \frac{1}{k}\,\big|\{\,i \in \text{top-}k(w)\;:\;\mathrm{cat}(i)=c\,\}\big| \in [0,1], \label{eq:purity}
\end{equation}

\begin{equation}
\textbf{Clustered row:}\quad
\mathbf{1}_{\text{cluster}}(w) = \mathbb{I}\!\big[p(w) \ge \tau\big], \quad \text{for threshold }\tau \in [0,1]. \label{eq:clustered}
\end{equation}

Here, $\mathcal{C}$ denotes the set of item categories, $\mathrm{cat}(i)$ returns the category of item $i$, and $\tau$ controls the strictness of cluster assignment. In simple terms, affinity quantifies the relative alignment of an FFN neuron’s value vector with item versus text embeddings, thereby indicating modality preference. Purity measures the concentration of a neuron’s top-$k$ nearest neighbors within a single item category, reflecting category-specific specialization. Clustered rows are those neurons whose purity exceeds a threshold $\tau$, identifying dimensions of the FFN value space that form coherent category-level clusters.

\section{Experiments}\label{sec:experiments}

\begin{table}[t]
\centering
\caption{Results on small Amazon catalogs. Highlight = LLM-Based. Bold = best; underline = second best; “–” = unreported. $^{1}$ \citet{zhai2025mm-quant-gen}. $^{2}$ \citet{cao2024gdm-aligning}. $^{3}$ \citet{zhang2025cove}.}
\resizebox{\textwidth}{!}{
\begin{tabular}{l *{12}{c}}
\toprule
\textbf{Method} &
\multicolumn{2}{c}{\textbf{Games}} &
\multicolumn{2}{c}{\textbf{Instruments}} &
\multicolumn{2}{c}{\textbf{Arts}} &
\multicolumn{2}{c}{\textbf{Sports}} &
\multicolumn{2}{c}{\textbf{Beauty}} &
\multicolumn{2}{c}{\textbf{Toys}} \\
\cmidrule(lr){2-3}\cmidrule(lr){4-5}\cmidrule(lr){6-7}\cmidrule(lr){8-9}\cmidrule(lr){10-11}\cmidrule(lr){12-13}
& NDCG@10 & HR@10
& NDCG@10 & HR@10
& NDCG@10 & HR@10
& NDCG@10 & HR@10
& NDCG@10 & HR@10
& NDCG@10 & HR@10 \\
\midrule
GRU4Rec$^{1,2}$   & 0.0453 & 0.0895 & 0.0857 & 0.1207 & 0.0690 & 0.1088 & 0.0110 & 0.0204 & 0.0137 & 0.0283 & 0.0084 & 0.0176 \\
Bert4Rec$^{1,2}$  & 0.0366 & 0.0725 & 0.0739 & 0.1081 & 0.0575 & 0.0922 & 0.0099 & 0.0191 & 0.0170 & 0.0347 & 0.0099 & 0.0203 \\
FDSA$^{1,2}$      & 0.0509 & 0.0988 & 0.0859 & 0.1249 & 0.0695 & 0.1190 & 0.0156 & 0.0288 & 0.0208 & 0.0407 & 0.0189 & 0.0381 \\
S3-Rec$^{1,2}$    & 0.0468 & 0.0903 & 0.0743 & 0.1123 & 0.0630 & 0.1030 & 0.0240 & 0.0385 & 0.0327 & 0.0647 & 0.0376 & 0.0700 \\
TIGER$^{1,2}$     & 0.0453 & 0.0857 & 0.0950 & 0.1221 & 0.0806 & 0.1167 & 0.0225 & 0.0400 & 0.0384 & 0.0648 & 0.0432 & 0.0712 \\
VQ-Rec$^1$    & 0.0329 & 0.0679 & 0.0891 & 0.1357 & 0.0844 & \underline{0.1386} & -& -& -& -& -& -\\
MISSRec$^1$   & 0.0499 & 0.1048 & 0.0880 & 0.1361 & 0.0815 & 0.1321 & -&- &- & -& -&- \\
\rowcolor{lightgray}P5-CID$^1$   & 0.0454 & 0.0824 & 0.0704 & 0.1119 & 0.0662 & 0.0994 &- &- & -& -& -&- \\
\rowcolor{lightgray}VIP5$^1$      & 0.0418 & 0.0758 & 0.0872 & 0.1071 & 0.0635 & 0.0859 &- & -& -& -& -& -\\ 
MQL4GRec$^1$  & 0.0548 & 0.1033 & \textbf{0.1060} & \underline{0.1375} & \underline{0.0950} & 0.1327 & -& -&- & -&- & -\\
\rowcolor{lightgray}ReAT$^2$      &- & -& -& -&- &- & 0.0232 & 0.0422 & 0.0535 & 0.0722 & 0.0461 & 0.0776 \\
\rowcolor{lightgray}E4SRec$^2$    & - &- &- & -&- & -& 0.0237 & 0.0410 & 0.0435 & 0.0758 & 0.0479 & 0.0798\\
IDGenRec$^2$  & - &- &- & -&- & -& \underline{0.0372} & 0.0574 & 0.0541 & 0.0814 & \underline{0.0551} & 0.0870\\ 
\rowcolor{lightgray}CoVE$^3$& - &- &- & -&- & -& 0.0359 & \underline{0.0624} & \underline{0.0593} & 0.1009 & \textbf{0.0595} & \textbf{0.0986} \\ \hline
SASRec    & 0.0547 & 0.0997 & 0.0749 & 0.1256 & 0.0927 & 0.1290 & 0.0289 & 0.0531 & 0.0541 & \underline{0.0945} & 0.0542 & \underline{0.0958}\\
HSTU      & \textbf{0.0609} & \underline{0.1089} & 0.0712 & 0.1214 & 0.0941 & 0.1301 & 0.0287 & 0.0515 & 0.0474 & 0.0863 & 0.0536 & 0.0933\\
ID Transformer    & 0.0392 & 0.0669 & 0.0709 & 0.0761 & 0.0824 & 0.1025 & 0.0081 & 0.0122 & 0.0314 & 0.0503 & 0.0271 & 0.0405\\
\rowcolor{lightgray}Text-Attr LLM & 0.0464 & 0.0862 & 0.0778 & 0.1133 & 0.0938 & 0.1374 & 0.0251 & 0.0497 & 0.0390 & 0.0761 & 0.0502 & 0.0895 \\ 
\rowcolor{lightgray}Item-LLM  & 0.0407 & 0.0734 & 0.0943 & 0.1095 & 0.0901 & 0.1272 & 0.0211 & 0.0369 & 0.0449 & 0.0738 & 0.0410 & 0.0704\\
\rowcolor{lightgray}\textbf{\abbr}      & \underline{0.0605} & \textbf{0.1102}& \underline{0.1054} & \textbf{0.1385} & \textbf{0.1029} & \textbf{0.1409} & \textbf{0.0391} & \textbf{0.0674} & \textbf{0.0665} & \textbf{0.1104} & 0.0531 & 0.0927 \\
\bottomrule
\end{tabular}
}
\label{tab:small_amazon_by_dataset}
\end{table}

\subsection{Experimental Settings}

\paragraph{Baselines}
Our main focus is on LLM-based recommenders, so the most relevant baselines are different ways of adding recommendation capability to LLMs. We include established LLM-for-Rec baselines that are directly comparable to our setting: the P5/P5-CID family, which reframes recommendation as text-to-text generation over a pretrained language model~\citep{geng2022RLP,hua2023p5-cid}; VIP5, a multimodal extension of P5 that adapts the LLM with parameter-efficient modules~\citep{geng2023vip5}; E4SRec, which keeps the LLM largely frozen and adds a lightweight ID-side adapter for sequential recommendation~\citep{li2023e4srec}; and ReAT, which aligns LLMs to recommendation objectives via auxiliary, recommendation-specific generated tasks~\citep{cao2024gdm-aligning}. These capture the main design choices for adding recommendation capability to LLMs (prompting, adapters, frozen-backbone adapters, alignment), and thus form our most relevant comparison set. In addition, we compare three variants built on the same backbone: (i) \emph{ID Transformer}, trained only on item tokens; (ii) \emph{Item-ID LLM + text-derived bias}~\citep{jiang2025urm}, where ID embeddings are augmented with text features; and (iii) \emph{Item-LLM}, which integrates item text via vocabulary expansion but without MoE. These three variants are matched to \abbr in parameter count and trained under identical token budgets.
For completeness, we also report results of classical sequence models (GRU4Rec~\citep{hidasi2015gru4rec}, Bert4Rec~\citep{sun2019bert4rec}, FDSA~\citep{zhang2019fdsa}, S3-Rec~\citep{zhou2020s3}), recent quantized/contrastive approaches (VQ-Rec~\citep{hou2023vq-req}, MissRec~\citep{wang2023missrec}, TIGER~\citep{rajput2023tiger}, MQL4GRec~\citep{zhai2025mm-quant-gen}, IDGenRec~\citep{tan2024idgenrec}), and strong transformer baselines (SASRec~\citep{kang2018sasrec}, HSTU~\citep{zhai2024hstu}). We further include CoVE~\citep{zhang2025cove}, which extends an LLM with LoRA parameters to encode catalog items. While these embedding-driven or classical models are not our primary comparison targets, we include them for completeness on smaller Amazon datasets. Full baseline details are in Appendix~\ref{app:baselines}.

\paragraph{Datasets, Evaluation, \& Backbone}
We use public Amazon Dataset: Games, Instruments and Arts~\citep{ni2019amzn} as well as Sports, Beauty and Toys~\cite{mcauley2015amzn}. We further report performance on larger 2023 Amazon variants (Beauty, Books, and Toys) with substantially larger item vocabularies~\cite{hou2024bridginglanguageitemsretrieval}. We also train and evaluate on our in-house industrial-scale dataset with hundreds of millions of users and tens of thousands of items. We report NDCG@10, HR@10 and MRR. Metrics are computed over the full catalog on Amazon datasets and on 50000 samples in our industrial dataset. We follow the standard leave last item out procedure for separating train and test datasets. All LLM-based models that we train, use \texttt{Qwen/Qwen2.5-0.5B} on text-analysis results, Amazon datasets, and for all ablations. We use \texttt{Qwen/Qwen2.5-1.5B} for main results on our proprietary dataset. See Appendix~\ref{sec:app-experiments} for all details.

\subsubsection{Results: Amazon Catalogs }
Table \ref{tab:small_amazon_by_dataset} summarizes performance across six small Amazon datasets. We observe that classical sequence models such as GRU4Rec~\citep{hidasi2015gru4rec} and Bert4Rec~\cite{sun2019bert4rec} perform consistently worse than more recent architectures, confirming the difficulty of modeling sparse item interactions in these settings. Transformer-based methods with additional inductive biases, such as FDSA~\citep{zhang2019fdsa}, S3-Rec~\cite{zhou2020s3}, and TIGER~\cite{rajput2023tiger}, provide moderate gains, while recent quantization and multi-modal approaches like VQ-Rec~\cite{hou2023vq-req}, MISSRec~\cite{wang2023missrec}, and MQL4GRec~\cite{zhai2025mm-quant-gen} achieve stronger results. Compared to direct LLM-Based baselines (highlighted in gray) and classical sequence models, \abbr delivers the most consistent improvements: it achieves the highest NDCG@10 and HR@10 in nearly all domains. These results highlight the robustness of our approach across diverse catalog sizes and domains, suggesting better generalization than prior methods that either overfit to specific datasets or fail to transfer across settings. 

We evaluate SASRec~\citep{kang2018sasrec}, HSTU~\citep{zhai2024hstu}, ID-Transformer, LLM-based baselines and \abbr on  Larger Amazon datasets. Table~\ref{tab:large_amazon_dataset_results} presents the results. \abbr is the strongest LLM-based method across all three catalogs: it is the top LLM on Beauty (2nd overall behind HSTU by a small margin), and it achieves the best overall scores on Books and Toys. In contrast, Item-LLM and Text-Attr LLM~\cite{jiang2025urm} lag behind \abbr across metrics, indicating that simply mixing item/text tokens or adding text-derived biases is insufficient. Together, these results support our claim that disentangling item and language pathways yields higher ranking quality than prior LLM baselines while remaining competitive with the best non-LLM models.

\begin{table}[t]
\centering
\begin{minipage}{0.49\linewidth}
\vspace{-20pt}
\centering
\captionof{table}{Results on large Amazon catalogs. Bold=best; underline=second best; Highlight=LLM-Based}
\vspace{-5pt}
\resizebox{\linewidth}{!}{
\begin{tabular}{l *{6}{c}}
\toprule
\textbf{Method} &
\multicolumn{2}{c}{\textbf{Beauty}} &
\multicolumn{2}{c}{\textbf{Books}}  &
\multicolumn{2}{c}{\textbf{Toys}} \\
\cmidrule(lr){2-3}\cmidrule(lr){4-5} \cmidrule(lr){6-7}
& NDCG@10  & HR@10
& NDCG@10  & HR@10
& NDCG@10  & HR@10\\
\midrule
SASRec         & 0.0051 & 0.0101 & 0.0064 & 0.0128 & 0.0122 & 0.0245\\
HSTU           & \textbf{0.0130} & \textbf{0.0247} & \underline{0.0211} & \underline{0.0410} & 0.0149 & 0.0332\\
ID Transformer & 0.0068 & 0.0095 &  \textbf{0.0224} & 0.0295 & 0.0048 & 0.0079 \\ 
\rowcolor{lightgray}Text-Attr LLM  & 0.0105 & 0.0163 & 0.0195 & 0.0290 & 0.0164 & 0.0300  \\
\rowcolor{lightgray}Item-LLM       & 0.0082 & 0.0119 & 0.0174 & 0.0261 & 0.0079 & 0.0148 \\
\rowcolor{lightgray} \abbr          & \underline{0.0119} & \underline{0.0228} & \textbf{0.0224} & \textbf{0.0419} & \textbf{0.0186} & \textbf{0.0361} \\
\bottomrule
\end{tabular}
\label{tab:large_amazon_dataset_results}
}
\end{minipage}
\hfill
\begin{minipage}{0.49\linewidth}
\centering
\vspace{-2pt}
\captionof{figure}{Results on our industrial dataset.}
\includegraphics[width=\linewidth]{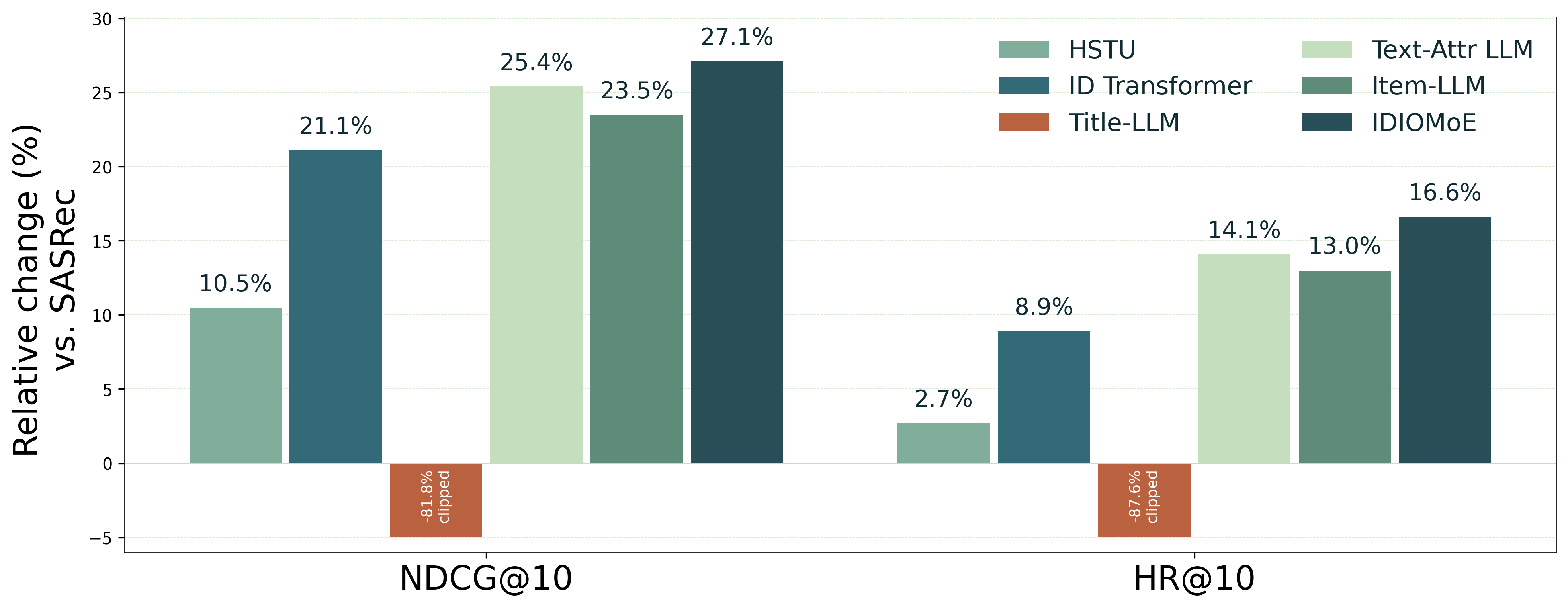}
\label{fig:propritary-results}
\end{minipage}
\end{table}

\subsubsection{Results: Proprietary Dataset}

While results on the Amazon datasets remain a useful reference point, we acknowledge their limitations. The benchmarks are relatively small and may contain overlaps that make them easier than real-world scenarios. Therefore, although we report results on these datasets for comparability with prior work, we place greater weight on evaluations conducted on our large-scale proprietary dataset, which we consider a more realistic and meaningful test of recommendation quality.

Figure~\ref{fig:propritary-results} (Table~\ref{tab:proprietary_by_dataset}) shows results on our large-scale proprietary dataset as improvements over the SASRec~\cite{kang2018sasrec} baseline. ID-Transformer achieves strong gains, confirming that transformers can effectively capture collaborative filtering signals when grounded in IDs and given enough compute. Title-LLM, which relies solely on free-form item titles, collapses in performance, highlighting the limitations of text-only representations for recommendation. Item-LLM combines IDs with textual features and yields further improvements, particularly on HR@10, demonstrating the value of jointly modeling collaborative and semantic signals. HSTU~\cite{zhai2024hstu} provides modest gains but falls short compared to the LLM-based approaches and doesn't support explainable recommendation. Finally, our method (\abbr) achieves the largest improvements across all metrics (+27.1\% NDCG@10, +16.6\% HR@10, +31.2\% MRR), showing that disentangling ID and text processing with specialized experts and routing not only preserves collaborative filtering strength but also better leverages semantic cues for robust large-scale recommendation.

\subsection{Ablations}

\subsubsection{Non-MoE capacity controls.}
To ensure that the improvements of \abbr are not simply due to added parameters, we compare against non-MoE variants with matched capacity.
Specifically, we consider three settings: (i) \emph{wide-FFN}, where the feed-forward layers of the transformer blocks are widened to match \abbr’s parameter count; (ii) \emph{append-blocks}, where additional transformer layers are added after the original stack; and (iii) \emph{prepend-blocks}, where extra layers are inserted before the original stack.
All models are trained under the same setup as \abbr with the hyperparameters and the same FLOPS. We also compare against a LoRA~\cite{hu2022lora} variant where low-rank adapters are added across all layers. Table~\ref{tab:nonmoe_capacity} summarizes the results. 

\begin{wrapfigure}{r}{0.5\textwidth}
\centering
\captionof{table}{Non-MoE capacity controls on Amazon-Beauty and Industrial datasets. 
All variants are matched to \abbr in parameter count. Results are shown as relative improvements over Item-LLM.}
\resizebox{\linewidth}{!}{
\begin{tabular}{lcccc}
\toprule
\multirow{2}{*}{\textbf{Method}} & 
\multicolumn{2}{c}{\textbf{Amazon-Beauty} $\Delta(\%)$} & 
\multicolumn{2}{c}{\textbf{Industrial} $\Delta(\%)$} \\
\cmidrule(lr){2-3}\cmidrule(lr){4-5}
& NDCG@10 & HR@10 & NDCG@10 & HR@10 \\
\midrule
Item-LLM (baseline) & \multicolumn{2}{c}{---} & \multicolumn{2}{c}{---} \\ \hline
LoRA-LLM           &  +21.5\%  & +7.9\%         & -79.1\%  & -76.3\%         \\ \hline
Wide-FFN           &  +27.0\%  & +24.9\%        &  +3.8\%  & +1.3\%         \\
Append-blocks      &  -87.8\%  & -90.3\%        &  -5.5\%  & -5.3\%         \\
Prepend-blocks     &  -97.2\%  & -95.9\%        &  -15.3\% & -16.2\%        \\ \hline
MoA                &  +48.3\%  & +46.2\%        &  +20.9\% & \underline{+27.1\%}        \\
MoT                &  \textbf{+49.3\%}  & \textbf{+51.1\%}        &  \underline{+22.5\%} & +24.8\%        \\
\abbr              &  \underline{+48.1\%}  & \underline{+49.6\%}        &  \textbf{+24.1\%} & \textbf{+28.9\%}        \\
\bottomrule
\end{tabular}
}
\label{tab:nonmoe_capacity}
\end{wrapfigure}

We find that simply adding parameters in non-structured ways is insufficient. Wide-FFN improves performance on Amazon-Beauty but only marginally helps in the industrial setting. In contrast, append-blocks and prepend-blocks severely degrade performance across both datasets, likely due to disruption of pretrained representations or training instability. LoRA-LLM, where low-rank adapters are added across all layers, helps slightly on Amazon-Beauty but fails drastically on the industrial benchmark, highlighting its sensitivity to scale and signal sparsity.

We also compare with various MoE designs. Both MoA (expert attention modules) and MoT (expertized full transformer blocks with cross attention) yield large improvements over all non-MoE controls. Importantly, \abbr performs on par or better than both, despite using a simpler and more efficient expert design focused solely on FFNs with static routing.  \rebuttal{Although MoA and MoT are competitive on Amazon-Beauty and occasionally match or slightly exceed \abbr there, we emphasize the industrial-scale results as our primary evidence. On this large setting, the FFN-based MoE of \abbr consistently outperforms MoA/MoT variants. Nonetheless, the pattern we observe might be dataset-dependent. The core idea of \abbr is to treat catalog items as first-class citizens and to separate where information about IDs and text is stored. All three MoE variants we ablate are consistent with this idea. Our choice to place MoE in the FFNs is guided by the stronger and more stable gains we see on the large-scale industrial dataset.}

These results confirm that \abbr's performance is not due to raw parameter count, but rather due to its intentional separation of item and language processing via token-type MoE routing. Compared to generic scaling or lightweight tuning (e.g., LoRA), the structured, disentangled pathways in \abbr yield higher accuracy, especially in large-scale settings where interference between item IDs and natural language is more pronounced.

\subsubsection{Item expert capacity}
We vary the intermediate width of the item expert per layer by applying different shrink factors to the middle layer of the item FFN experts. Larger shrink factors reduce the parameter count and latency, but they also constrain the model’s ability to capture rich collaborative signals. Table~\ref{tab:shrink_ablation} presents the results.
On Amazon-Beauty, we see that moderate shrink values (2 and 4) provide substantial improvements over the baseline, with shrink=4 yielding the best balance of capacity and efficiency (+41.8\% NDCG@10, +26.6\% HR@10). However, very aggressive shrinking (shrink=8) reduces gains, suggesting that the item expert becomes under-parameterized.In contrast, results on the industrial dataset show a different trend: shrinking consistently hurts performance, with small but steady drops in both NDCG@10 and HR@10 as capacity decreases. 
\begin{wrapfigure}{r}{0.5\linewidth}
\centering
\captionof{table}{Impact of varying item expert capacity.}
\resizebox{\linewidth}{!}{
\begin{tabular}{lcccc}
\toprule
\multirow{2}{*}{\textbf{Shrink}} & 
\multicolumn{2}{c}{\textbf{Amazon-Beauty} $\Delta(\%)$} & 
\multicolumn{2}{c}{\textbf{Industrial} $\Delta(\%)$} \\
\cmidrule(lr){2-3}\cmidrule(lr){4-5}
& NDCG@10 & HR@10 & NDCG@10 & HR@10 \\
\midrule
1 (baseline) & \multicolumn{2}{c}{---} & \multicolumn{2}{c}{\textbf{---}}    \\
2            &   +21.5\%     &  +23.3\%      &  -2.0\%      &  -2.1\%     \\
4            &   \textbf{+41.8\%}     &  \textbf{+26.6\%}      &  -3.1\%      &  -2.2\%     \\
8            &   +10.1\%     &  +6.6\%       &  -4.5\%      &  -3.6\%     \\
\bottomrule
\end{tabular}
}
\label{tab:shrink_ablation}
\vspace{-20pt}
\end{wrapfigure}
 These findings indicate that while smaller benchmarks can benefit from lighter experts, large-scale real-world data demands higher item-expert capacity to preserve recommendation accuracy. This motivates the need for adaptive capacity allocation, where expert width can be tuned to match the complexity and scale of the target domain. Our method provides this control on capacity allocation.

\subsubsection{Where to Insert MoE Layers}
To study where MoE layers are most effective, we conduct an ablation by selecting different insertion strategies. Specifically, we activate MoE experts in (i) the first 8 layers, (ii) the middle 8 layers, (iii) the last 8 layers, and (iv) every third layer throughout the model. This allows us to compare the impact of placing MoE capacity in shallow, intermediate, deep, or evenly distributed positions. We report results on the Amazon-Arts dataset in Table~\ref{tab:moe_layer_ablation}.

\begin{wrapfigure}{r}{0.5\textwidth}
\centering
\captionof{table}{Ablation on where to insert MoE layers.}
\resizebox{\linewidth}{!}{
\begin{tabular}{lcccc}
\toprule
\multirow{2}{*}{\textbf{MoE Placement}} & \multicolumn{2}{c}{\textbf{Amazon-Beauty} $\Delta(\%)$} &
\multicolumn{2}{c}{\textbf{Industrial} $\Delta(\%)$} \\
\cmidrule(lr){2-3}\cmidrule(lr){4-5}
& NDCG@10 & HR@10 & NDCG@10 & HR@10 \\
\midrule
First 8 (baseline) & \multicolumn{2}{c}{---} & \multicolumn{2}{c}{---} \\
Every 3            & +17.7\% & +10.3\% & +2.0\%   & +5.3\% \\
Middle 8           & +22.8\% & +17.2\% & +3.1\%   & +6.9\% \\
Last 8             & \textbf{+28.4\%} & \textbf{+27.6\%} & \textbf{+9.6\%}   & \textbf{+9.0\%} \\
\bottomrule
\end{tabular}
\vspace{-10pt}
}
\label{tab:moe_layer_ablation}
\end{wrapfigure}

We observe clear differences depending on where MoE layers are inserted. Using MoE in the first 8 layers yields the weakest performance, suggesting that early representations are dominated by low-level token processing where additional capacity is less beneficial. Distributing MoE every three layers achieves moderate improvements but still falls short. Placing MoE in the middle 8 layers improves results, but the largest gains come from inserting MoE in the last 8 layers (+27.6\% HR@10 and +28.4\% NDCG@10 over baseline). This indicates that deeper layers (where task-specific semantics and collaborative filtering patterns are most prominent) benefit most from specialized experts, as they directly shape the final ranking representations.

\subsubsection{Static vs. Dynamic Routing}
\begin{wrapfigure}{r}{0.5\linewidth}
\centering
\captionof{table}{Impact of static routing.}
\resizebox{\linewidth}{!}{
\begin{tabular}{lcccc}
\toprule
\multirow{2}{*}{\textbf{Routing Strategy}} & 
\multicolumn{2}{c}{\textbf{Amazon-Beauty} $\Delta(\%)$} & 
\multicolumn{2}{c}{\textbf{Industrial} $\Delta(\%)$} \\
\cmidrule(lr){2-3}\cmidrule(lr){4-5}
& NDCG@10 & HR@10 & NDCG@10 & HR@10 \\
\midrule
Static & \multicolumn{2}{c}{\textbf{---}} & \multicolumn{2}{c}{\textbf{---}}    \\
Dynamic            &   -59.5\%     &  -36.9\%      &  -24.2\%      &  -24.4\%     \\
\bottomrule
\end{tabular}
}
\label{tab:routing_ablation}
\end{wrapfigure}

We find that a switch-style~\citep{fedus2022switch} dynamic gating severely degrades recommendation quality, while static token-type routing performs much better (Table \ref{tab:shrink_ablation}). The likely reason is that static routing gives each expert a clear, consistent role (language vs. item IDs) so they can specialize without interference. In contrast, dynamic routing mixes assignments across experts, leading to greater entanglement between signals and weaker specialization. This highlights that a fixed separation by token type is not just simpler but also more effective for disentangling language and recommendation signals.

For each layer $\ell$, we report means/medians of $a(w)$~(Equation~\ref{eq:affinity}) and $p(w)$~(Equation~\ref{eq:purity}) across rows, and the \emph{clustered fraction} $\mathbb{E}[\mathbf{1}_{\text{cluster}}(w)]$~(Equation~\ref{eq:clustered}). In MoE, we compare the item expert. We extract $W_{\text{out}}$ rows, compute top-$k$ similarities to items and text, and summarize per layer and overall. We set $k{=}20$ and $\tau{=}0.5$.

\subsection{FFN Key-Value Memory Analysis}

\begin{figure}[t]
    \centering
    \begin{subfigure}[b]{0.48\textwidth}
        \centering
        \includegraphics[width=\linewidth]{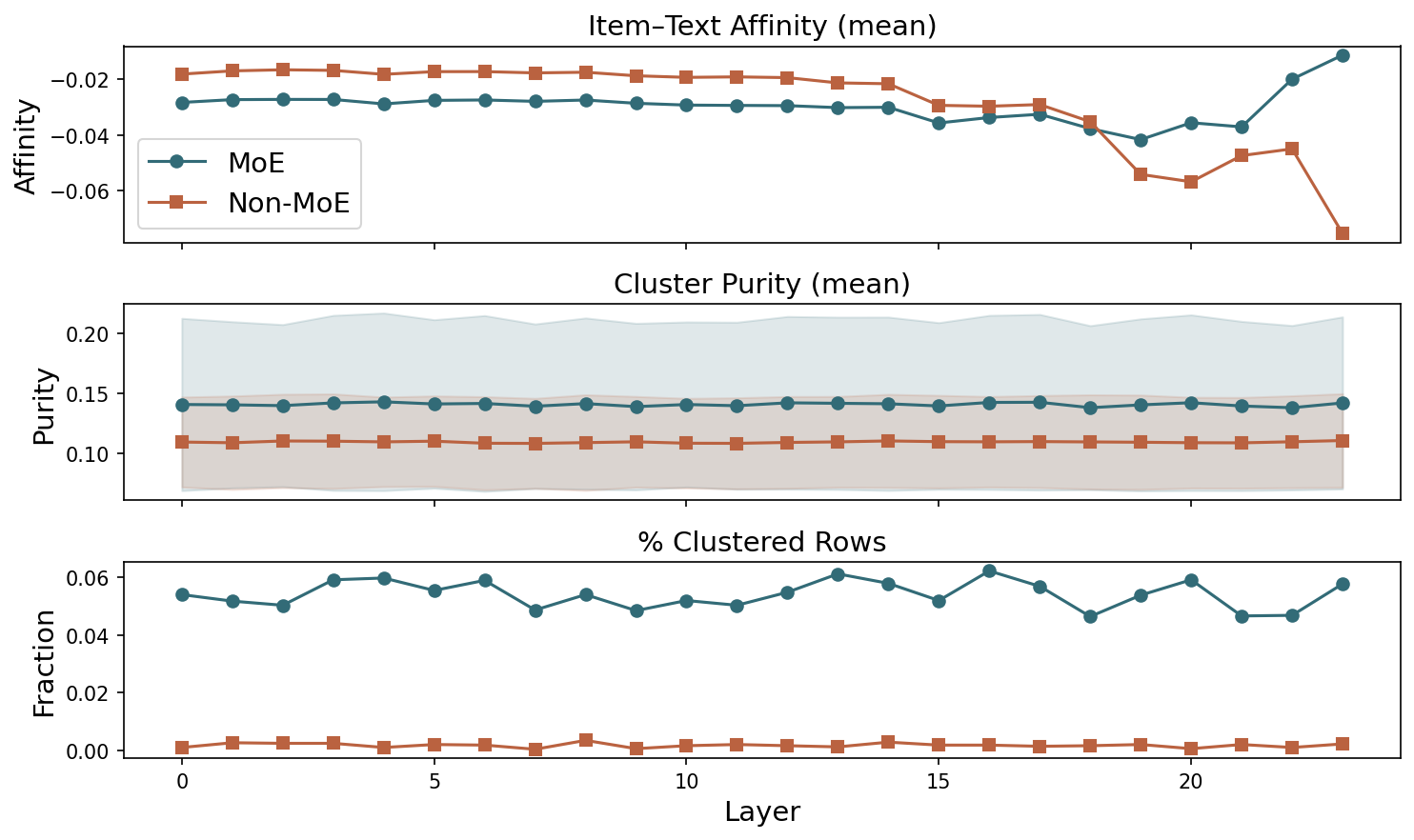}
        \caption{Amazon-Arts}
        \label{fig:ffn_analysis_amazon}
    \end{subfigure}
    \hfill
    \begin{subfigure}[b]{0.48\textwidth}
        \centering
        \includegraphics[width=\linewidth]{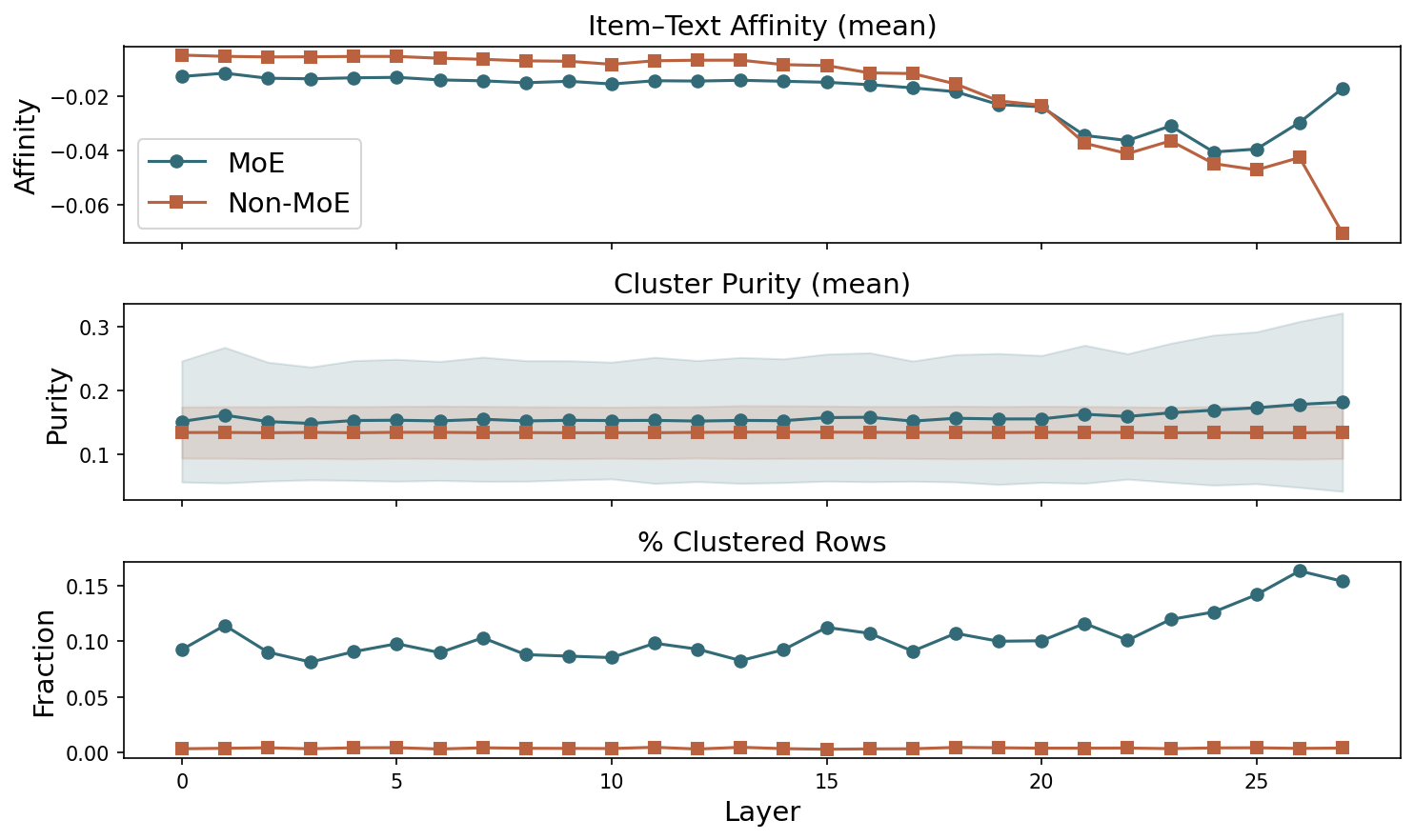}
        \caption{Industrial dataset}
        \label{fig:ffn_analysis_industrial}
    \end{subfigure}
    \caption{FFN key-value memory analysis comparing MoE vs. non-MoE. 
    Each subfigure shows item-text affinity, cluster purity, and fraction of clustered rows across transformer layers.}
    \label{fig:ffn_kv_analysis}
\end{figure}

The results in Figure~\ref{fig:ffn_kv_analysis} show clear differences between MoE and non-MoE models when analyzing FFN neurons as key-value memories. In terms of item-text affinity, both models begin with weak modality preference, but deeper layers of the non-MoE baseline drift toward negative affinity (favoring text), whereas the MoE model maintains more balanced alignment. This indicates that MoE preserves item sensitivity in upper layers, where recommendation decisions are most critical ~(Table~\ref{tab:moe_layer_ablation}).  

For cluster purity, MoE consistently yields higher values across layers, meaning that its neurons are more category-specific: when a neuron activates for items, it tends to retrieve items from the same category. Similarly, the fraction of clustered rows~(neurons forming coherent category-level clusters) remains low and flat for the non-MoE baseline, is always higher in MoE and rises sharply in the later layers of MoE on the more challenging industrial dataset. Together, these results suggest that MoE separation leads to clearer item-text specialization, higher category purity, and more structured clustering than a vanilla transformer, reinforcing our claim that expert separation enables more interpretable and modular representations of recommendation signals.

\section{Conclusion}\label{sec:conclusion}
We introduced \abbr, a dual-expert continued-pretrained language model that processes text and item data through two specialized experts. Despite its simplicity, \abbr outperforms both classical and recently proposed LLM-based recommendation models. It effectively preserves the pretrained knowledge of the LLM.  Our findings highlight the importance of using specialized sub-networks for different modalities, rather than scaling indiscriminately with a single model for all inputs. We view \abbr as a step toward more sustainable and adaptive LLMs for recommendation tasks, and believe this direction is crucial in our efforts to achieve better recommendation performance and interpretability without relying on unnecessarily large models that exhibit diminishing returns.

\section{Acknowledgments}\label{sec:ack}
The authors would like to thank Zahra Miri for her assistance in preparing the figures.\\
This work was partially supported by NSF IIS 2347592, 2348169, DBI 2405416, CCF 2348306, CNS 2347617, RISE 2536663.

\bibliography{references}

@String(CVPR  = {IEEE Conf. Comput. Vis. Pattern Recog.})

@String(NeurIPS = {Adv. Neural Inform. Process. Syst.})

@String(ICLR  = {Int. Conf. Learn. Represent.})

@String(AAAI  = {AAAI})

@String(IJCAI = {IJCAI})

@String(CVPR  = {CVPR})

@String(NeurIPS = {NeurIPS})

@String(ICLR  = {ICLR})

@Inbook{Lops2011content,
author="Lops, Pasquale
and de Gemmis, Marco
and Semeraro, Giovanni",
title="Content-based Recommender Systems: State of the Art and Trends",
bookTitle="Recommender Systems Handbook",
year="2011",
publisher="Springer US",
address="Boston, MA",
pages="73--105",
isbn="978-0-387-85820-3",
}

@article{koren2009matrix,
  title={Matrix factorization techniques for recommender systems},
  author={Koren, Yehuda and Bell, Robert and Volinsky, Chris},
  journal={Computer},
  volume={42},
  number={8},
  pages={30--37},
  year={2009},
  publisher={IEEE}
}

@article{cao2024gdm-aligning,
  title={Aligning large language models with recommendation knowledge},
  author={Cao, Yuwei and Mehta, Nikhil and Yi, Xinyang and Keshavan, Raghunandan and Heldt, Lukasz and Hong, Lichan and Chi, Ed H and Sathiamoorthy, Maheswaran},
  journal={arXiv preprint arXiv:2404.00245},
  year={2024}
}

@article{jiang2025urm,
  title={Large Language Model as Universal Retriever in Industrial-Scale Recommender System},
  author={Jiang, Junguang and Huang, Yanwen and Liu, Bin and Kong, Xiaoyu and Li, Xinhang and Xu, Ziru and Zhu, Han and Xu, Jian and Zheng, Bo},
  journal={arXiv preprint arXiv:2502.03041},
  year={2025}
}

@article{zhang2025cove,
  title={CoVE: Compressed Vocabulary Expansion Makes Better LLM-based Recommender Systems},
  author={Zhang, Haochen and Zhang, Tianyi and Yin, Junze and Gal, Oren and Shrivastava, Anshumali and Braverman, Vladimir},
  journal={arXiv preprint arXiv:2506.19993},
  year={2025}
}

@article{zhai2025mm-quant-gen,
  title={Multimodal Quantitative Language for Generative Recommendation},
  author={Zhai, Jianyang and Mai, Zi-Feng and Wang, Chang-Dong and Yang, Feidiao and Zheng, Xiawu and Li, Hui and Tian, Yonghong},
  journal={arXiv preprint arXiv:2504.05314},
  year={2025}
}

@article{abdollahpouri2019managing-pop-bias,
  title={Managing popularity bias in recommender systems with personalized re-ranking},
  author={Abdollahpouri, Himan and Burke, Robin and Mobasher, Bamshad},
  journal={arXiv preprint arXiv:1901.07555},
  year={2019}
}

@article{yang2025gr-llm,
  title={GR-LLMs: Recent Advances in Generative Recommendation Based on Large Language Models},
  author={Yang, Zhen and Lin, Haitao and Zhang, Ziji and others},
  journal={arXiv preprint arXiv:2507.06507},
  year={2025}
}

@inproceedings{kang2018sasrec,
  title={Self-attentive sequential recommendation},
  author={Kang, Wang-Cheng and McAuley, Julian},
  booktitle={2018 IEEE international conference on data mining (ICDM)},
  pages={197--206},
  year={2018},
  organization={IEEE}
}

@inproceedings{geng2022RLP,
  title={Recommendation as language processing (rlp): A unified pretrain, personalized prompt \& predict paradigm (p5)},
  author={Geng, Shijie and Liu, Shuchang and Fu, Zuohui and Ge, Yingqiang and Zhang, Yongfeng},
  booktitle={Proceedings of the 16th ACM conference on recommender systems},
  pages={299--315},
  year={2022}
}

@inproceedings{sun2019bert4rec,
  title={BERT4Rec: Sequential recommendation with bidirectional encoder representations from transformer},
  author={Sun, Fei and Liu, Jun and Wu, Jian and Pei, Changhua and Lin, Xiao and Ou, Wenwu and Jiang, Peng},
  booktitle={Proceedings of the 28th ACM international conference on information and knowledge management},
  pages={1441--1450},
  year={2019}
}

@article{hidasi2015gru4rec,
  title={Session-based recommendations with recurrent neural networks},
  author={Hidasi, Bal{\'a}zs and Karatzoglou, Alexandros and Baltrunas, Linas and Tikk, Domonkos},
  journal={arXiv preprint arXiv:1511.06939},
  year={2015}
}

@inproceedings{hou2024llm-zeroshot-ranker,
  title={Large language models are zero-shot rankers for recommender systems},
  author={Hou, Yupeng and Zhang, Junjie and Lin, Zihan and Lu, Hongyu and Xie, Ruobing and McAuley, Julian and Zhao, Wayne Xin},
  booktitle={European Conference on Information Retrieval},
  pages={364--381},
  year={2024},
  organization={Springer}
}

@article{lu2024aligning-llm-rl,
  title={Aligning large language models for controllable recommendations},
  author={Lu, Wensheng and Lian, Jianxun and Zhang, Wei and Li, Guanghua and Zhou, Mingyang and Liao, Hao and Xie, Xing},
  journal={arXiv preprint arXiv:2403.05063},
  year={2024}
}

@inproceedings{zhu2024ccllm4rec,
title={Collaborative Large Language Model for Recommender Systems},
author={Yaochen Zhu and Liang Wu and Qi Guo and Liangjie Hong and Jundong Li},
booktitle={The Web Conference 2024},
year={2024},
}

@article{zhai2024hstu,
  title={Actions speak louder than words: Trillion-parameter sequential transducers for generative recommendations},
  author={Zhai, Jiaqi and Liao, Lucy and Liu, Xing and Wang, Yueming and Li, Rui and Cao, Xuan and Gao, Leon and Gong, Zhaojie and Gu, Fangda and He, Michael and others},
  journal={arXiv preprint arXiv:2402.17152},
  year={2024}
}

@article{rajput2023tiger,
  title={Recommender systems with generative retrieval},
  author={Rajput, Shashank and Mehta, Nikhil and Singh, Anima and Hulikal Keshavan, Raghunandan and Vu, Trung and Heldt, Lukasz and Hong, Lichan and Tay, Yi and Tran, Vinh and Samost, Jonah and others},
  journal={Advances in Neural Information Processing Systems},
  volume={36},
  pages={10299--10315},
  year={2023}
}

@article{deng2025onerec,
  title={Onerec: Unifying retrieve and rank with generative recommender and iterative preference alignment},
  author={Deng, Jiaxin and Wang, Shiyao and Cai, Kuo and Ren, Lejian and Hu, Qigen and Ding, Weifeng and Luo, Qiang and Zhou, Guorui},
  journal={arXiv preprint arXiv:2502.18965},
  year={2025}
}

@article{han2025mtgr,
  title={MTGR: Industrial-Scale Generative Recommendation Framework in Meituan},
  author={Han, Ruidong and Yin, Bin and Chen, Shangyu and Jiang, He and Jiang, Fei and Li, Xiang and Ma, Chi and Huang, Mincong and Li, Xiaoguang and Jing, Chunzhen and others},
  journal={arXiv preprint arXiv:2505.18654},
  year={2025}
}

@misc{shazeer2017outrageously,
    title   = {Outrageously Large Neural Networks: The Sparsely-Gated Mixture-of-Experts Layer},
    author  = {Noam Shazeer and Azalia Mirhoseini and Krzysztof Maziarz and Andy Davis and Quoc Le and Geoffrey Hinton and Jeff Dean},
    year    = {2017},
    eprint  = {1701.06538},
    archivePrefix = {arXiv},
    primaryClass = {cs.LG}
}

@misc{lepikhin2020gshard,
    title   = {GShard: Scaling Giant Models with Conditional Computation and Automatic Sharding},
    author  = {Dmitry Lepikhin and HyoukJoong Lee and Yuanzhong Xu and Dehao Chen and Orhan Firat and Yanping Huang and Maxim Krikun and Noam Shazeer and Zhifeng Chen},
    year    = {2020},
    eprint  = {2006.16668},
    archivePrefix = {arXiv},
    primaryClass = {cs.CL}
}

@article{fedus2022switch,
  title={Switch transformers: Scaling to trillion parameter models with simple and efficient sparsity},
  author={Fedus, William and Zoph, Barret and Shazeer, Noam},
  journal={Journal of Machine Learning Research},
  volume={23},
  number={120},
  pages={1--39},
  year={2022}
}

@misc{qwen2025qwen25technicalreport,
      title={Qwen2.5 Technical Report}, 
      author={Qwen and : and An Yang and Baosong Yang and Beichen Zhang and Binyuan Hui and Bo Zheng and Bowen Yu and Chengyuan Li and Dayiheng Liu and Fei Huang and Haoran Wei and Huan Lin and Jian Yang and Jianhong Tu and Jianwei Zhang and Jianxin Yang and Jiaxi Yang and Jingren Zhou and Junyang Lin and Kai Dang and Keming Lu and Keqin Bao and Kexin Yang and Le Yu and Mei Li and Mingfeng Xue and Pei Zhang and Qin Zhu and Rui Men and Runji Lin and Tianhao Li and Tianyi Tang and Tingyu Xia and Xingzhang Ren and Xuancheng Ren and Yang Fan and Yang Su and Yichang Zhang and Yu Wan and Yuqiong Liu and Zeyu Cui and Zhenru Zhang and Zihan Qiu},
      year={2025},
      eprint={2412.15115},
      archivePrefix={arXiv},
      primaryClass={cs.CL},
}

@inproceedings{ni2019amzn,
  title={Justifying recommendations using distantly-labeled reviews and fine-grained aspects},
  author={Ni, Jianmo and Li, Jiacheng and McAuley, Julian},
  booktitle={Proceedings of the 2019 conference on empirical methods in natural language processing and the 9th international joint conference on natural language processing (EMNLP-IJCNLP)},
  pages={188--197},
  year={2019}
}

@inproceedings{mcauley2015amzn,
  title={Image-based recommendations on styles and substitutes},
  author={McAuley, Julian and Targett, Christopher and Shi, Qinfeng and Van Den Hengel, Anton},
  booktitle={Proceedings of the 38th international ACM SIGIR conference on research and development in information retrieval},
  pages={43--52},
  year={2015}
}

@misc{amazon_rufus2024,
  title   = {Amazon's Rufus AI assistant now available to all U.S. customers},
  author  = {{Amazon}},
  year    = {2024},
  url     = {https://www.aboutamazon.com/news/retail/how-to-use-amazon-rufus},
  note    = {Accessed 2025-08-28}
}

@misc{meta_llama3_2024,
  title   = {Meet Your New Assistant: Meta AI, Built with Llama 3},
  author  = {{Meta}},
  year    = {2024},
  url     = {https://about.fb.com/news/2024/04/meta-ai-assistant-built-with-llama-3/},
  note    = {Accessed 2025-08-28}
}

@misc{netflix_fm_2025,
  title   = {Foundation Model for Personalized Recommendation},
  author  = {Netflix},
  year    = {2025},
  url     = {https://netflixtechblog.com/foundation-model-for-personalized-recommendation-1a0bd8e02d39},
  note    = {Accessed 2025-08-28}
}

@misc{zhu2025collaborativeretrievallargelanguagenetflix,
      title={Collaborative Retrieval for Large Language Model-based Conversational Recommender Systems}, 
      author={Yaochen Zhu and Chao Wan and Harald Steck and Dawen Liang and Yesu Feng and Nathan Kallus and Jundong Li},
      year={2025},
      eprint={2502.14137},
      archivePrefix={arXiv},
      primaryClass={cs.IR},
}

@inproceedings{zhang2019fdsa,
  title={Feature-level deeper self-attention network for sequential recommendation.},
  author={Zhang, Tingting and Zhao, Pengpeng and Liu, Yanchi and Sheng, Victor S and Xu, Jiajie and Wang, Deqing and Liu, Guanfeng and Zhou, Xiaofang and others},
  booktitle={IJCAI},
  pages={4320--4326},
  year={2019}
}

@inproceedings{zhou2020s3,
  title={S3-rec: Self-supervised learning for sequential recommendation with mutual information maximization},
  author={Zhou, Kun and Wang, Hui and Zhao, Wayne Xin and Zhu, Yutao and Wang, Sirui and Zhang, Fuzheng and Wang, Zhongyuan and Wen, Ji-Rong},
  booktitle={Proceedings of the 29th ACM international conference on information \& knowledge management},
  pages={1893--1902},
  year={2020}
}

@inproceedings{hou2023vq-req,
  title={Learning vector-quantized item representation for transferable sequential recommenders},
  author={Hou, Yupeng and He, Zhankui and McAuley, Julian and Zhao, Wayne Xin},
  booktitle={Proceedings of the ACM Web Conference 2023},
  pages={1162--1171},
  year={2023}
}

@inproceedings{wang2023missrec,
  title={Missrec: Pre-training and transferring multi-modal interest-aware sequence representation for recommendation},
  author={Wang, Jinpeng and Zeng, Ziyun and Wang, Yunxiao and Wang, Yuting and Lu, Xingyu and Li, Tianxiang and Yuan, Jun and Zhang, Rui and Zheng, Hai-Tao and Xia, Shu-Tao},
  booktitle={Proceedings of the 31st ACM International Conference on Multimedia},
  pages={6548--6557},
  year={2023}
}

@inproceedings{hua2023p5-cid,
  title={How to index item ids for recommendation foundation models},
  author={Hua, Wenyue and Xu, Shuyuan and Ge, Yingqiang and Zhang, Yongfeng},
  booktitle={Proceedings of the Annual International ACM SIGIR Conference on Research and Development in Information Retrieval in the Asia Pacific Region},
  pages={195--204},
  year={2023}
}

@article{geng2023vip5,
  title={Vip5: Towards multimodal foundation models for recommendation},
  author={Geng, Shijie and Tan, Juntao and Liu, Shuchang and Fu, Zuohui and Zhang, Yongfeng},
  journal={arXiv preprint arXiv:2305.14302},
  year={2023}
}

@article{li2023e4srec,
  title={E4srec: An elegant effective efficient extensible solution of large language models for sequential recommendation},
  author={Li, Xinhang and Chen, Chong and Zhao, Xiangyu and Zhang, Yong and Xing, Chunxiao},
  journal={arXiv preprint arXiv:2312.02443},
  year={2023}
}

@inproceedings{tan2024idgenrec,
  title={Idgenrec: Llm-recsys alignment with textual id learning},
  author={Tan, Juntao and Xu, Shuyuan and Hua, Wenyue and Ge, Yingqiang and Li, Zelong and Zhang, Yongfeng},
  booktitle={Proceedings of the 47th international ACM SIGIR conference on research and development in information retrieval},
  year={2024}
}

@article{geva2022key-val-memory-analysis,
  title={Transformer feed-forward layers build predictions by promoting concepts in the vocabulary space},
  author={Geva, Mor and Caciularu, Avi and Wang, Kevin Ro and Goldberg, Yoav},
  journal={arXiv preprint arXiv:2203.14680},
  year={2022}
}

@misc{merity2016wikitext,
      title={Pointer Sentinel Mixture Models},
      author={Stephen Merity and Caiming Xiong and James Bradbury and Richard Socher},
      year={2016},
      eprint={1609.07843},
      archivePrefix={arXiv},
      primaryClass={cs.CL}
}

@inproceedings{hu2022lora,
title={Lo{RA}: Low-Rank Adaptation of Large Language Models},
author={Edward J Hu and yelong shen and Phillip Wallis and Zeyuan Allen-Zhu and Yuanzhi Li and Shean Wang and Lu Wang and Weizhu Chen},
booktitle={International Conference on Learning Representations},
year={2022},
}

@article{suzgun2022bbh,
  title={Challenging BIG-Bench Tasks and Whether Chain-of-Thought Can Solve Them},
  author={Suzgun, Mirac and Scales, Nathan and Sch{\"a}rli, Nathanael and Gehrmann, Sebastian and Tay, Yi and Chung, Hyung Won and Chowdhery, Aakanksha and Le, Quoc V and Chi, Ed H and Zhou, Denny and and Wei, Jason},
  journal={arXiv preprint arXiv:2210.09261},
  year={2022}
}

@inproceedings{zellers2019hellaswag,
    title={HellaSwag: Can a Machine Really Finish Your Sentence?},
    author={Zellers, Rowan and Holtzman, Ari and Bisk, Yonatan and Farhadi, Ali and Choi, Yejin},
    booktitle ={Proceedings of the 57th Annual Meeting of the Association for Computational Linguistics},
    year={2019}
}

@article{hendryckstest2021mmlu,
      title={Measuring Massive Multitask Language Understanding},
      author={Dan Hendrycks and Collin Burns and Steven Basart and Andy Zou and Mantas Mazeika and Dawn Song and Jacob Steinhardt},
      journal={Proceedings of the International Conference on Learning Representations (ICLR)},
      year={2021}
}

@misc{sakaguchi2019winogrande,
      title={WinoGrande: An Adversarial Winograd Schema Challenge at Scale}, 
      author={Keisuke Sakaguchi and Ronan Le Bras and Chandra Bhagavatula and Yejin Choi},
      year={2019},
      eprint={1907.10641},
      archivePrefix={arXiv},
      primaryClass={cs.CL},
}

@misc{hou2024bridginglanguageitemsretrieval,
      title={Bridging Language and Items for Retrieval and Recommendation}, 
      author={Yupeng Hou and Jiacheng Li and Zhankui He and An Yan and Xiusi Chen and Julian McAuley},
      year={2024},
      eprint={2403.03952},
      archivePrefix={arXiv},
      primaryClass={cs.IR},
}

@misc{eval-harness,
  author       = {Gao, Leo and Tow, Jonathan and Abbasi, Baber and Biderman, Stella and Black, Sid and DiPofi, Anthony and Foster, Charles and Golding, Laurence and Hsu, Jeffrey and Le Noac'h, Alain and Li, Haonan and McDonell, Kyle and Muennighoff, Niklas and Ociepa, Chris and Phang, Jason and Reynolds, Laria and Schoelkopf, Hailey and Skowron, Aviya and Sutawika, Lintang and Tang, Eric and Thite, Anish and Wang, Ben and Wang, Kevin and Zou, Andy},
  title        = {The Language Model Evaluation Harness},
  month        = 07,
  year         = 2024,
  publisher    = {Zenodo},
  version      = {v0.4.3},
}

@misc{zeng2020knowledgetransferpretrainingrecommendation,
      title={Knowledge Transfer via Pre-training for Recommendation: A Review and Prospect}, 
      author={Zheni Zeng and Chaojun Xiao and Yuan Yao and Ruobing Xie and Zhiyuan Liu and Fen Lin and Leyu Lin and Maosong Sun},
      year={2020},
      eprint={2009.09226},
      archivePrefix={arXiv},
      primaryClass={cs.IR},
}

@misc{liu2023pretrainpromptrecommendationcomprehensive,
      title={Pre-train, Prompt and Recommendation: A Comprehensive Survey of Language Modelling Paradigm Adaptations in Recommender Systems}, 
      author={Peng Liu and Lemei Zhang and Jon Atle Gulla},
      year={2023},
      eprint={2302.03735},
      archivePrefix={arXiv},
      primaryClass={cs.IR},
}

@misc{lin2024recommendersystemsbenefitlarge,
      title={How Can Recommender Systems Benefit from Large Language Models: A Survey}, 
      author={Jianghao Lin and Xinyi Dai and Yunjia Xi and Weiwen Liu and Bo Chen and Hao Zhang and Yong Liu and Chuhan Wu and Xiangyang Li and Chenxu Zhu and Huifeng Guo and Yong Yu and Ruiming Tang and Weinan Zhang},
      year={2024},
      eprint={2306.05817},
      archivePrefix={arXiv},
      primaryClass={cs.IR},
}

@misc{yuan2023recommendersystemsidvs,
      title={Where to Go Next for Recommender Systems? ID- vs. Modality-based Recommender Models Revisited}, 
      author={Zheng Yuan and Fajie Yuan and Yu Song and Youhua Li and Junchen Fu and Fei Yang and Yunzhu Pan and Yongxin Ni},
      year={2023},
      eprint={2303.13835},
      archivePrefix={arXiv},
      primaryClass={cs.IR},
}

@misc{wang2024generativerecommendationnextgenerationrecommender,
      title={Generative Recommendation: Towards Next-generation Recommender Paradigm}, 
      author={Wenjie Wang and Xinyu Lin and Fuli Feng and Xiangnan He and Tat-Seng Chua},
      year={2024},
      eprint={2304.03516},
      archivePrefix={arXiv},
      primaryClass={cs.IR},
}

@inproceedings{Fu_2024_adapter,
   title={Exploring Adapter-based Transfer Learning for Recommender Systems: Empirical Studies and Practical Insights},
   booktitle={Proceedings of the 17th ACM International Conference on Web Search and Data Mining},
   publisher={ACM},
   author={Fu, Junchen and Yuan, Fajie and Song, Yu and Yuan, Zheng and Cheng, Mingyue and Cheng, Shenghui and Zhang, Jiaqi and Wang, Jie and Pan, Yunzhu},
   year={2024},
}

@misc{yuan2018simpleconvolutionalgenerativenetwork,
      title={A Simple Convolutional Generative Network for Next Item Recommendation}, 
      author={Fajie Yuan and Alexandros Karatzoglou and Ioannis Arapakis and Joemon M Jose and Xiangnan He},
      year={2018},
      eprint={1808.05163},
      archivePrefix={arXiv},
      primaryClass={cs.IR},
}

@inproceedings{transformer4rec,
author = {de Souza Pereira Moreira, Gabriel and Rabhi, Sara and Lee, Jeong Min and Ak, Ronay and Oldridge, Even},
title = {Transformers4Rec: Bridging the Gap between NLP and Sequential / Session-Based Recommendation},
year = {2021},
publisher = {Association for Computing Machinery},
booktitle = {Proceedings of the 15th ACM Conference on Recommender Systems},
}

@misc{hou2022universalsequencerepresentationlearning,
      title={Towards Universal Sequence Representation Learning for Recommender Systems}, 
      author={Yupeng Hou and Shanlei Mu and Wayne Xin Zhao and Yaliang Li and Bolin Ding and Ji-Rong Wen},
      year={2022},
      eprint={2206.05941},
      archivePrefix={arXiv},
      primaryClass={cs.IR},
}

@misc{hou2023learningvectorquantizeditemrepresentation,
      title={Learning Vector-Quantized Item Representation for Transferable Sequential Recommenders}, 
      author={Yupeng Hou and Zhankui He and Julian McAuley and Wayne Xin Zhao},
      year={2023},
      eprint={2210.12316},
      archivePrefix={arXiv},
      primaryClass={cs.IR},
}

@misc{yuan2020parameterefficienttransfersequentialbehaviors,
      title={Parameter-Efficient Transfer from Sequential Behaviors for User Modeling and Recommendation}, 
      author={Fajie Yuan and Xiangnan He and Alexandros Karatzoglou and Liguang Zhang},
      year={2020},
      eprint={2001.04253},
      archivePrefix={arXiv},
      primaryClass={cs.IR},
}

@misc{xiao2021uprecuserawarepretrainingrecommender,
      title={UPRec: User-Aware Pre-training for Recommender Systems}, 
      author={Chaojun Xiao and Ruobing Xie and Yuan Yao and Zhiyuan Liu and Maosong Sun and Xu Zhang and Leyu Lin},
      year={2021},
      eprint={2102.10989},
      archivePrefix={arXiv},
      primaryClass={cs.IR},
}

@inproceedings{qiu2021u,
  title={U-BERT: Pre-training user representations for improved recommendation},
  author={Qiu, Zhaopeng and Wu, Xian and Gao, Jingyue and Fan, Wei},
  booktitle={Proceedings of the AAAI Conference on Artificial Intelligence},
  volume={35},
  year={2021}
}

@misc{
li2021userbert,
title={User{\{}BERT{\}}: Self-supervised User Representation Learning},
author={Tianyu Li and Ali Cevahir and Derek Cho and Hao Gong and DuyKhuong Nguyen and Bjorn Stenger},
year={2021},
}

@misc{yuan2021personmodelworldlearning,
      title={One Person, One Model, One World: Learning Continual User Representation without Forgetting}, 
      author={Fajie Yuan and Guoxiao Zhang and Alexandros Karatzoglou and Joemon Jose and Beibei Kong and Yudong Li},
      year={2021},
      eprint={2009.13724},
      archivePrefix={arXiv},
      primaryClass={cs.IR},
}

@misc{shin2022scalinglawrecommendationmodels,
      title={Scaling Law for Recommendation Models: Towards General-purpose User Representations}, 
      author={Kyuyong Shin and Hanock Kwak and Su Young Kim and Max Nihlen Ramstrom and Jisu Jeong and Jung-Woo Ha and Kyung-Min Kim},
      year={2022},
      eprint={2111.11294},
      archivePrefix={arXiv},
      primaryClass={cs.IR},
}

@misc{yao2021selfsupervisedlearninglargescaleitem,
      title={Self-supervised Learning for Large-scale Item Recommendations}, 
      author={Tiansheng Yao and Xinyang Yi and Derek Zhiyuan Cheng and Felix Yu and Ting Chen and Aditya Menon and Lichan Hong and Ed H. Chi and Steve Tjoa and Jieqi Kang and Evan Ettinger},
      year={2021},
      eprint={2007.12865},
      archivePrefix={arXiv},
      primaryClass={cs.LG},
}

@misc{wang2025transreclearningtransferablerecommendation,
      title={TransRec: Learning Transferable Recommendation from Mixture-of-Modality Feedback}, 
      author={Jie Wang and Fajie Yuan and Mingyue Cheng and Joemon M. Jose and Chenyun Yu and Beibei Kong and Zhijin Wang and Bo Hu and Zang Li},
      year={2025},
      eprint={2206.06190},
      archivePrefix={arXiv},
      primaryClass={cs.IR},
}

@misc{li2022inttowergenerationtwotowermodel,
      title={IntTower: the Next Generation of Two-Tower Model for Pre-Ranking System}, 
      author={Xiangyang Li and Bo Chen and HuiFeng Guo and Jingjie Li and Chenxu Zhu and Xiang Long and Sujian Li and Yichao Wang and Wei Guo and Longxia Mao and Jinxing Liu and Zhenhua Dong and Ruiming Tang},
      year={2022},
      eprint={2210.09890},
      archivePrefix={arXiv},
      primaryClass={cs.IR},
}

@misc{he2015vbprvisualbayesianpersonalized,
      title={VBPR: Visual Bayesian Personalized Ranking from Implicit Feedback}, 
      author={Ruining He and Julian McAuley},
      year={2015},
      eprint={1510.01784},
      archivePrefix={arXiv},
      primaryClass={cs.IR},
}

@inproceedings{
zhang2021language,
title={Language Models as Recommender Systems: Evaluations and Limitations},
author={Yuhui Zhang and HAO DING and Zeren Shui and Yifei Ma and James Zou and Anoop Deoras and Hao Wang},
booktitle={I (Still) Can't Believe It's Not Better! NeurIPS 2021 Workshop},
year={2021},
}

@inproceedings{muhamed2021ctr,
  title={CTR-BERT: Cost-effective knowledge distillation for billion-parameter teacher models},
  author={Muhamed, Aashiq and Keivanloo, Iman and Perera, Sujan and Mracek, James and Xu, Yi and Cui, Qingjun and Rajagopalan, Santosh and Zeng, Belinda and Chilimbi, Trishul},
  booktitle={NeurIPS Efficient Natural Language and Speech Processing Workshop},
  year={2021}
}

@misc{cui2022m6recgenerativepretrainedlanguage,
      title={M6-Rec: Generative Pretrained Language Models are Open-Ended Recommender Systems}, 
      author={Zeyu Cui and Jianxin Ma and Chang Zhou and Jingren Zhou and Hongxia Yang},
      year={2022},
      eprint={2205.08084},
      archivePrefix={arXiv},
      primaryClass={cs.IR},
}

@misc{liu2022ptabusingpretrainedlanguage,
      title={PTab: Using the Pre-trained Language Model for Modeling Tabular Data}, 
      author={Guang Liu and Jie Yang and Ledell Wu},
      year={2022},
      eprint={2209.08060},
      archivePrefix={arXiv},
      primaryClass={cs.LG},
}

@inproceedings{Zhang_2023,
   title={Prompt Learning for News Recommendation},
   booktitle={Proceedings of the 46th International ACM SIGIR Conference on Research and Development in Information Retrieval},
   publisher={ACM},
   author={Zhang, Zizhuo and Wang, Bang},
   year={2023},}

@misc{wei2024llmreclargelanguagemodels,
      title={LLMRec: Large Language Models with Graph Augmentation for Recommendation}, 
      author={Wei Wei and Xubin Ren and Jiabin Tang and Qinyong Wang and Lixin Su and Suqi Cheng and Junfeng Wang and Dawei Yin and Chao Huang},
      year={2024},
      eprint={2311.00423},
      archivePrefix={arXiv},
      primaryClass={cs.IR},
}

@misc{liu2023chatgptgoodrecommenderpreliminary,
      title={Is ChatGPT a Good Recommender? A Preliminary Study}, 
      author={Junling Liu and Chao Liu and Peilin Zhou and Renjie Lv and Kang Zhou and Yan Zhang},
      year={2023},
      eprint={2304.10149},
      archivePrefix={arXiv},
      primaryClass={cs.IR},
}

@inproceedings{Dai_2023,
   title={Uncovering ChatGPT’s Capabilities in Recommender Systems},
   booktitle={Proceedings of the 17th ACM Conference on Recommender Systems},
   publisher={ACM},
   author={Dai, Sunhao and Shao, Ninglu and Zhao, Haiyuan and Yu, Weijie and Si, Zihua and Xu, Chen and Sun, Zhongxiang and Zhang, Xiao and Xu, Jun},
   year={2023},}

@misc{lin2023sparksartificialgeneralrecommender,
      title={Sparks of Artificial General Recommender (AGR): Early Experiments with ChatGPT}, 
      author={Guo Lin and Yongfeng Zhang},
      year={2023},
      eprint={2305.04518},
      archivePrefix={arXiv},
      primaryClass={cs.IR},
}

@inproceedings{Bao_2023,
   title={TALLRec: An Effective and Efficient Tuning Framework to Align Large Language Model with Recommendation},
   booktitle={Proceedings of the 17th ACM Conference on Recommender Systems},
   publisher={ACM},
   author={Bao, Keqin and Zhang, Jizhi and Zhang, Yang and Wang, Wenjie and Feng, Fuli and He, Xiangnan},
   year={2023},}

@misc{yang2023palrpersonalizationawarellms,
      title={PALR: Personalization Aware LLMs for Recommendation}, 
      author={Fan Yang and Zheng Chen and Ziyan Jiang and Eunah Cho and Xiaojiang Huang and Yanbin Lu},
      year={2023},
      eprint={2305.07622},
      archivePrefix={arXiv},
      primaryClass={cs.IR},
}

@misc{zhang2023recommendationinstructionfollowinglarge,
      title={Recommendation as Instruction Following: A Large Language Model Empowered Recommendation Approach}, 
      author={Junjie Zhang and Ruobing Xie and Yupeng Hou and Wayne Xin Zhao and Leyu Lin and Ji-Rong Wen},
      year={2023},
      eprint={2305.07001},
      archivePrefix={arXiv},
      primaryClass={cs.IR},
}

@misc{friedman2023leveraginglargelanguagemodels,
      title={Leveraging Large Language Models in Conversational Recommender Systems}, 
      author={Luke Friedman and Sameer Ahuja and David Allen and Zhenning Tan and Hakim Sidahmed and Changbo Long and Jun Xie and Gabriel Schubiner and Ajay Patel and Harsh Lara and Brian Chu and Zexi Chen and Manoj Tiwari},
      year={2023},
      eprint={2305.07961},
      archivePrefix={arXiv},
      primaryClass={cs.IR},
}

@misc{carranza2024syntheticquerygenerationprivacypreserving,
      title={Synthetic Query Generation for Privacy-Preserving Deep Retrieval Systems using Differentially Private Language Models}, 
      author={Aldo Gael Carranza and Rezsa Farahani and Natalia Ponomareva and Alex Kurakin and Matthew Jagielski and Milad Nasr},
      year={2024},
      eprint={2305.05973},
      archivePrefix={arXiv},
      primaryClass={cs.CL},
}

@misc{li2023exploringupperlimitstextbased,
      title={Exploring the Upper Limits of Text-Based Collaborative Filtering Using Large Language Models: Discoveries and Insights}, 
      author={Ruyu Li and Wenhao Deng and Yu Cheng and Zheng Yuan and Jiaqi Zhang and Fajie Yuan},
      year={2023},
      eprint={2305.11700},
      archivePrefix={arXiv},
      primaryClass={cs.IR},
}

@misc{bao2023bistepgroundingparadigmlarge,
      title={A Bi-Step Grounding Paradigm for Large Language Models in Recommendation Systems}, 
      author={Keqin Bao and Jizhi Zhang and Wenjie Wang and Yang Zhang and Zhengyi Yang and Yancheng Luo and Chong Chen and Fuli Feng and Qi Tian},
      year={2023},
      eprint={2308.08434},
      archivePrefix={arXiv},
      primaryClass={cs.IR},
}

@misc{li2023ctrlconnectcollaborativelanguage,
      title={CTRL: Connect Collaborative and Language Model for CTR Prediction}, 
      author={Xiangyang Li and Bo Chen and Lu Hou and Ruiming Tang},
      year={2023},
      eprint={2306.02841},
      archivePrefix={arXiv},
      primaryClass={cs.IR},
}

@misc{yue2023llamarectwostagerecommendationusing,
      title={LlamaRec: Two-Stage Recommendation using Large Language Models for Ranking}, 
      author={Zhenrui Yue and Sara Rabhi and Gabriel de Souza Pereira Moreira and Dong Wang and Even Oldridge},
      year={2023},
      eprint={2311.02089},
      archivePrefix={arXiv},
      primaryClass={cs.IR},
}

@misc{wang2023llm4visexplainablevisualizationrecommendation,
      title={LLM4Vis: Explainable Visualization Recommendation using ChatGPT}, 
      author={Lei Wang and Songheng Zhang and Yun Wang and Ee-Peng Lim and Yong Wang},
      year={2023},
      eprint={2310.07652},
      archivePrefix={arXiv},
      primaryClass={cs.HC},
}

@inproceedings{ma2018modeling,
  title={Modeling task relationships in multi-task learning with multi-gate mixture-of-experts},
  author={Ma, Jiaqi and Zhao, Zhe and Yi, Xinyang and Chen, Jilin and Hong, Lichan and Chi, Ed H},
  booktitle={Proceedings of the 24th ACM SIGKDD international conference on knowledge discovery \& data mining},
  pages={1930--1939},
  year={2018}
}

@inproceedings{moe-rec-2,
author = {Tang, Hongyan and Liu, Junning and Zhao, Ming and Gong, Xudong},
title = {Progressive Layered Extraction (PLE): A Novel Multi-Task Learning (MTL) Model for Personalized Recommendations},
year = {2020},
publisher = {Association for Computing Machinery},
booktitle = {Proceedings of the 14th ACM Conference on Recommender Systems},
}

@inproceedings{mome,
  author={Jiahui Xu and Lu Sun and Dengji Zhao},
  title={MoME: Mixture-of-Masked-Experts for Efficient Multi-Task Recommendation},
  year={2024},
  cdate={1704067200000},
  pages={2527-2531},
  booktitle={SIGIR},

}

@inproceedings{Zhang_2024moe,
   title={M3oE: Multi-Domain Multi-Task Mixture-of Experts Recommendation Framework},
   booktitle={Proceedings of the 47th International ACM SIGIR Conference on Research and Development in Information Retrieval},
   publisher={ACM},
   author={Zhang, Zijian and Liu, Shuchang and Yu, Jiaao and Cai, Qingpeng and Zhao, Xiangyu and Zhang, Chunxu and Liu, Ziru and Liu, Qidong and Zhao, Hongwei and Hu, Lantao and Jiang, Peng and Gai, Kun},
   year={2024},
}

@misc{wang2024homehierarchymultigateexperts,
      title={HoME: Hierarchy of Multi-Gate Experts for Multi-Task Learning at Kuaishou}, 
      author={Xu Wang and Jiangxia Cao and Zhiyi Fu and Kun Gai and Guorui Zhou},
      year={2024},
      eprint={2408.05430},
      archivePrefix={arXiv},
      primaryClass={cs.IR}, 
}

@misc{li2019deepconversationalrecommendations,
      title={Towards Deep Conversational Recommendations}, 
      author={Raymond Li and Samira Kahou and Hannes Schulz and Vincent Michalski and Laurent Charlin and Chris Pal},
      year={2019},
      eprint={1812.07617},
      archivePrefix={arXiv},
      primaryClass={cs.LG},
}

@misc{chen2019knowledgebasedrecommenderdialog,
      title={Towards Knowledge-Based Recommender Dialog System}, 
      author={Qibin Chen and Junyang Lin and Yichang Zhang and Ming Ding and Yukuo Cen and Hongxia Yang and Jie Tang},
      year={2019},
      eprint={1908.05391},
      archivePrefix={arXiv},
      primaryClass={cs.CL},
}

@inproceedings{Kemper_2024,
   title={Retrieval-Augmented Conversational Recommendation with Prompt-based Semi-Structured Natural Language State Tracking},
   booktitle={Proceedings of the 47th International ACM SIGIR Conference on Research and Development in Information Retrieval},
   publisher={ACM},
   author={Kemper, Sara and Cui, Justin and Dicarlantonio, Kai and Lin, Kathy and Tang, Danjie and Korikov, Anton and Sanner, Scott},
   year={2024},}

@article{li2023gpt4rec,
  title={GPT4Rec: A generative framework for personalized recommendation and user interests interpretation},
  author={Li, Jinming and Zhang, Wentao and Wang, Tian and Xiong, Guanglei and Lu, Alan and Medioni, Gerard},
  journal={arXiv preprint arXiv:2304.03879},
  year={2023}
}

@article{seqreq-programming-pretrained,
author = {Tang, Min and Cui, Shujie and Jin, Zhe and Liang, Shiuan-ni and Li, Chenliang and Zou, Lixin},
title = {Sequential recommendation by reprogramming pretrained transformer},
year = {2025},
issue_date = {Jan 2025},
publisher = {Pergamon Press, Inc.},
address = {USA},
volume = {62},
number = {1},
issn = {0306-4573},
journal = {Inf. Process. Manage.},
month = jan,
numpages = {15}
}

@misc{ren2024easyrecsimpleeffectivelanguage,
      title={EasyRec: Simple yet Effective Language Models for Recommendation}, 
      author={Xubin Ren and Chao Huang},
      year={2024},
      eprint={2408.08821},
      archivePrefix={arXiv},
      primaryClass={cs.IR},
}

@article{zhao2025simaug,
  title={SimAug: Enhancing Recommendation with Pretrained Language Models for Dense and Balanced Data Augmentation},
  author={Zhao, Yuying and Yang, Xiaodong and Chen, Huiyuan and Fan, Xiran and Wang, Yu and Cai, Yiwei and Derr, Tyler},
  journal={arXiv preprint arXiv:2505.01695},
  year={2025}
}

@article{calrec,
  author = {Li, Yaoyiran and Zhai, Xiang and Alzantot, Moustafa and Yu, Keyi and Vulić, Ivan and Korhonen, Anna and Hammad, Mohamed},
  
  title = {CALRec: Contrastive Alignment of Generative LLMs for Sequential Recommendation},
  
  journal = {arXiv},
  
  year = {2024},
  
  copyright = {Creative Commons Attribution 4.0 International}
}

@inproceedings{Wang_2021-stackrec,
   title={StackRec: Efficient Training of Very Deep Sequential Recommender Models by Iterative Stacking},
   booktitle={Proceedings of the 44th International ACM SIGIR Conference on Research and Development in Information Retrieval},
   publisher={ACM},
   author={Wang, Jiachun and Yuan, Fajie and Chen, Jian and Wu, Qingyao and Yang, Min and Sun, Yang and Zhang, Guoxiao},
   year={2021},}

@misc{kieu2025keyworddrivenretrievalaugmentedlargelanguage,
      title={Keyword-driven Retrieval-Augmented Large Language Models for Cold-start User Recommendations}, 
      author={Hai-Dang Kieu and Minh Duc Nguyen and Thanh-Son Nguyen and Dung D. Le},
      year={2025},
      eprint={2405.19612},
      archivePrefix={arXiv},
      primaryClass={cs.IR},
}

@inproceedings{lu2023bert,
  title={BERT-RS: A neural personalized recommender system with BERT},
  author={Lu, Kezhi and Zhang, Qian and Zhang, Guangquan and Lu, Jie},
  booktitle={Machine Learning, Multi Agent and Cyber Physical Systems: Proceedings of the 15th International FLINS Conference (FLINS 2022)},
  pages={390--397},
  year={2023},
  organization={World Scientific}
}

@inproceedings{zhang2021unbert,
  title={UNBERT: User-News Matching BERT for News Recommendation.},
  author={Zhang, Qi and Li, Jingjie and Jia, Qinglin and Wang, Chuyuan and Zhu, Jieming and Wang, Zhaowei and He, Xiuqiang},
  booktitle={IJCAI},
  volume={21},
  pages={3356--3362},
  year={2021}
}

@misc{liang2025taxonomyguidedzeroshotrecommendationsllms,
      title={Taxonomy-Guided Zero-Shot Recommendations with LLMs}, 
      author={Yueqing Liang and Liangwei Yang and Chen Wang and Xiongxiao Xu and Philip S. Yu and Kai Shu},
      year={2025},
      eprint={2406.14043},
      archivePrefix={arXiv},
      primaryClass={cs.IR},
}

@misc{wu2024smalllanguagemodelsserve,
      title={Could Small Language Models Serve as Recommenders? Towards Data-centric Cold-start Recommendations}, 
      author={Xuansheng Wu and Huachi Zhou and Yucheng Shi and Wenlin Yao and Xiao Huang and Ninghao Liu},
      year={2024},
      eprint={2306.17256},
      archivePrefix={arXiv},
      primaryClass={cs.IR},
}

@inproceedings{Li_2024,
   title={Large Language Models for Next Point-of-Interest Recommendation},
   booktitle={Proceedings of the 47th International ACM SIGIR Conference on Research and Development in Information Retrieval},
   publisher={ACM},
   author={Li, Peibo and de Rijke, Maarten and Xue, Hao and Ao, Shuang and Song, Yang and Salim, Flora D.},
   year={2024}
}

@misc{contal2024ragsysitemcoldstartrecommenderrag,
      title={RAGSys: Item-Cold-Start Recommender as RAG System}, 
      author={Emile Contal and Garrin McGoldrick},
      year={2024},
      eprint={2405.17587},
      archivePrefix={arXiv},
      primaryClass={cs.IR},
}

@article{liu2023llava,
  title={Visual instruction tuning},
  author={Liu, Haotian and Li, Chunyuan and Wu, Qingyang and Lee, Yong Jae},
  journal={Advances in neural information processing systems},
  volume={36},
  pages={34892--34916},
  year={2023}
}

@article{lin2024moellava,
  title={Moe-llava: Mixture of experts for large vision-language models},
  author={Lin, Bin and Tang, Zhenyu and Ye, Yang and Cui, Jiaxi and Zhu, Bin and Jin, Peng and Huang, Jinfa and Zhang, Junwu and Pang, Yatian and Ning, Munan and others},
  journal={arXiv preprint arXiv:2401.15947},
  year={2024}
}

@article{li2025unimoe,
  title={Uni-moe: Scaling unified multimodal llms with mixture of experts},
  author={Li, Yunxin and Jiang, Shenyuan and Hu, Baotian and Wang, Longyue and Zhong, Wanqi and Luo, Wenhan and Ma, Lin and Zhang, Min},
  journal={IEEE Transactions on Pattern Analysis and Machine Intelligence},
  year={2025},
  publisher={IEEE}
}

@misc{diao2025eve2,
      title={EVEv2: Improved Baselines for Encoder-Free Vision-Language Models}, 
      author={Haiwen Diao and Xiaotong Li and Yufeng Cui and Yueze Wang and Haoge Deng and Ting Pan and Wenxuan Wang and Huchuan Lu and Xinlong Wang},
      year={2025},
      eprint={2502.06788},
      archivePrefix={arXiv},
      primaryClass={cs.CV},
      url={https://arxiv.org/abs/2502.06788}, 
}

@misc{deng2025bagel,
      title={Emerging Properties in Unified Multimodal Pretraining}, 
      author={Chaorui Deng and Deyao Zhu and Kunchang Li and Chenhui Gou and Feng Li and Zeyu Wang and Shu Zhong and Weihao Yu and Xiaonan Nie and Ziang Song and Guang Shi and Haoqi Fan},
      year={2025},
      eprint={2505.14683},
      archivePrefix={arXiv},
      primaryClass={cs.CV},
      url={https://arxiv.org/abs/2505.14683}, 
}

@misc{shen2023scalingvlmmoe,
      title={Scaling Vision-Language Models with Sparse Mixture of Experts}, 
      author={Sheng Shen and Zhewei Yao and Chunyuan Li and Trevor Darrell and Kurt Keutzer and Yuxiong He},
      year={2023},
      eprint={2303.07226},
      archivePrefix={arXiv},
      primaryClass={cs.CV},
      url={https://arxiv.org/abs/2303.07226}, 
}

@misc{bao2022vlmounifiedvlp,
      title={VLMo: Unified Vision-Language Pre-Training with Mixture-of-Modality-Experts}, 
      author={Hangbo Bao and Wenhui Wang and Li Dong and Qiang Liu and Owais Khan Mohammed and Kriti Aggarwal and Subhojit Som and Furu Wei},
      year={2022},
      eprint={2111.02358},
      archivePrefix={arXiv},
      primaryClass={cs.CV},
      url={https://arxiv.org/abs/2111.02358}, 
}

@inproceedings{xiong2025llava,
  title={Llava-critic: Learning to evaluate multimodal models},
  author={Xiong, Tianyi and Wang, Xiyao and Guo, Dong and Ye, Qinghao and Fan, Haoqi and Gu, Quanquan and Huang, Heng and Li, Chunyuan},
  booktitle={Proceedings of the Computer Vision and Pattern Recognition Conference},
  pages={13618--13628},
  year={2025}
}

@inproceedings{ganjdaneshnot,
  title={Not All Prompts Are Made Equal: Prompt-based Pruning of Text-to-Image Diffusion Models},
  author={Ganjdanesh, Alireza and Shirkavand, Reza and Gao, Shangqian and Huang, Heng},
  booktitle={The Thirteenth International Conference on Learning Representations},
  year={2025}
}

@inproceedings{shirkavandcost,
  title={Cost-Aware Contrastive Routing for LLMs},
  author={Shirkavand, Reza and Gao, Shangqian and Yu, Peiran and Huang, Heng},
  booktitle={The Thirty-ninth Annual Conference on Neural Information Processing Systems},
  year={2025}
}

@InProceedings{Shirkavand_2025_CVPR,
    author    = {Shirkavand, Reza and Yu, Peiran and Gao, Shangqian and Somepalli, Gowthami and Goldstein, Tom and Huang, Heng},
    title     = {Efficient Fine-Tuning and Concept Suppression for Pruned Diffusion Models},
    booktitle = {Proceedings of the IEEE/CVF Conference on Computer Vision and Pattern Recognition (CVPR)},
    month     = {June},
    year      = {2025},
    pages     = {18619-18629}
}

@inproceedings{shirkavandbilevel,
  title={Bilevel ZOFO: Efficient LLM Fine-Tuning and Meta-Training},
  author={Shirkavand, Reza and Yu, Peiran and He, Qi and Huang, Heng},
  booktitle={The Thirty-ninth Annual Conference on Neural Information Processing Systems},
  year={2025}
}

@inproceedings{rawal2025argus,
  title={Argus: Hallucination and omission evaluation in video-llms},
  author={Rawal, Ruchit and Shirkavand, Reza and Huang, Heng and Somepalli, Gowthami and Goldstein, Tom},
  booktitle={Proceedings of the IEEE/CVF International Conference on Computer Vision},
  pages={20280--20290},
  year={2025}
}

@article{xiong2026phycritic,
  title={PhyCritic: Multimodal Critic Models for Physical AI},
  author={Xiong, Tianyi and Wang, Shihao and Liu, Guilin and Dong, Yi and Li, Ming and Huang, Heng and Kautz, Jan and Yu, Zhiding},
  journal={arXiv preprint arXiv:2602.11124},
  year={2026}
}

@inproceedings{wang2025damo,
title={Damo: Decoding by accumulating activations momentum for mitigating hallucinations in vision-language models},
author={Wang, Kaishen and Gu, Hengrui and Gao, Meijun and Zhou, Kaixiong},
booktitle={The Thirteenth International Conference on Learning Representations},
year={2025}
}
\bibliographystyle{iclr2026_conference}

\newpage
\appendix
\section{Extended Related Work}\label{sec:related-work-full}

\subsection{Classic Recommendation Approaches}
Recommender systems have long relied on two complementary paradigms: collaborative filtering (CF)~\citep{yao2021selfsupervisedlearninglargescaleitem,wang2025transreclearningtransferablerecommendation,li2022inttowergenerationtwotowermodel,he2015vbprvisualbayesianpersonalized} and content-based (CB) methods.  
CF models exploit user–item interaction patterns, such as ratings or clicks, to learn latent representations of users and items~\citep{koren2009matrix}. This approach is domain-agnostic and often yields high accuracy, but it suffers from well-known \emph{cold-start} problems for new users or items and can exhibit strong popularity bias~\citep{abdollahpouri2019managing-pop-bias}, over-recommending popular items at the expense of long-tail discovery.  
CB methods instead leverage explicit item features or descriptions to recommend similar items, which can address item cold-start but ignore collaborative patterns and the “wisdom of the crowd.” These methods may produce over-specialized recommendations that limit serendipity.

Hybrid recommenders attempt to combine CF and CB to balance relevance, novelty, and coverage. However, even hybrid systems can be difficult to control with respect to multi-objective goals like fairness, diversity, or novelty without post-hoc re-ranking.

\paragraph{Sequential and Contextual Models.}  
Moving beyond static recommendation, sequential models~\citep{yuan2018simpleconvolutionalgenerativenetwork,zhou2020s3,transformer4rec,hou2022universalsequencerepresentationlearning,hou2023learningvectorquantizeditemrepresentation,wang2023missrec} predict a user’s next interaction by modeling temporal dependencies in their history. Early neural solutions include GRU4Rec~\citep{hidasi2015gru4rec}, which applied gated recurrent units to capture sequence dynamics. The introduction of Transformers brought a step-change: SASRec~\citep{kang2018sasrec} was the first to model next-item prediction in an autoregressive fashion using self-attention, improving short-term preference modeling. BERT4Rec~\citep{sun2019bert4rec} adapted bidirectional Transformers to better utilize context on both sides of a target position.  
These architectures form strong baselines in academic and industrial settings, yet they still rely on abstract IDs or dense embeddings, making it hard to integrate external semantic knowledge or to directly optimize multiple objectives beyond accuracy.

Recent work also explores fairness- and diversity-aware training, multi-objective loss formulations, and contextual augmentation, but these methods often require complex pipelines and lack the natural flexibility of a language interface.

\subsection{Large Language Models for Recommendation}
The advent of large language models (LLMs) pretrained~\citep{yuan2020parameterefficienttransfersequentialbehaviors,xiao2021uprecuserawarepretrainingrecommender,qiu2021u,li2021userbert,yuan2021personmodelworldlearning,shin2022scalinglawrecommendationmodels,shirkavandcost} on massive corpora has opened new opportunities for recommendation.~\citep{zeng2020knowledgetransferpretrainingrecommendation,liu2023pretrainpromptrecommendationcomprehensive,lin2024recommendersystemsbenefitlarge,yuan2023recommendersystemsidvs,wang2024generativerecommendationnextgenerationrecommender,Fu_2024_adapter}
LLMs provide broad world knowledge, reasoning skills, and instruction-following~\cite{zhang2023recommendationinstructionfollowinglarge,Li_2024,contal2024ragsysitemcoldstartrecommenderrag} abilities that can extend beyond the pattern-matching of traditional recommenders~\citep{zhang2021language,muhamed2021ctr,cui2022m6recgenerativepretrainedlanguage,liu2022ptabusingpretrainedlanguage,Zhang_2023,wei2024llmreclargelanguagemodels,li2023exploringupperlimitstextbased,wang2023llm4visexplainablevisualizationrecommendation}.

\paragraph{LLMs as Recommenders.}  
A pioneering example is P5~\citep{geng2022RLP}, which reformulates diverse recommendation tasks into a unified text-to-text format, allowing zero-shot~\cite{hou2024llm-zeroshot-ranker} and few-shot transfer between tasks such as rating prediction, sequential recommendation, and explanation generation~\citep{bao2023bistepgroundingparadigmlarge,li2023ctrlconnectcollaborativelanguage,yue2023llamarectwostagerecommendationusing,lu2023bert,zhang2021unbert,wu2024smalllanguagemodelsserve}. This unification facilitates integration of multiple modalities, such as textual descriptions or reviews, and enables natural-language queries~\cite{liu2023chatgptgoodrecommenderpreliminary,Bao_2023,Dai_2023,lin2023sparksartificialgeneralrecommender,Zhang_2023,yang2023palrpersonalizationawarellms,carranza2024syntheticquerygenerationprivacypreserving,kieu2025keyworddrivenretrievalaugmentedlargelanguage}.  
However, item representation in such setups is often token-inefficient—especially for large catalogs—because items must be described in text, and off-the-shelf LLMs lack direct exposure to collaborative signals from user–item interactions~\citep{cao2024gdm-aligning}. This leads to a mismatch between the LLM’s pretrained knowledge and the domain-specific collaborative knowledge needed for effective recommendation.

\paragraph{Zero-Shot and Prompt-Based Approaches.}  
Zero-shot prompting~\citep{hou2024llm-zeroshot-ranker,liang2025taxonomyguidedzeroshotrecommendationsllms} evaluates an LLM as a ranker given a user’s history and a set of candidate items in the prompt. Such methods can achieve competitive performance without task-specific training, demonstrating strong generalization, but are sensitive to prompt design, prone to sequence-order biases, and often ignore subtle interaction semantics.

\paragraph{Fine-Tuning and Alignment.}  
To address these limitations, fine-tuning methods adapt LLMs to recommendation tasks while preserving language capabilities~\cite{ren2024easyrecsimpleeffectivelanguage,zhao2025simaug,calrec,Wang_2021-stackrec,shirkavandbilevel}.  
GDM~\citep{cao2024gdm-aligning} introduces auxiliary natural-language training tasks (e.g., masked item modeling, BPR) to inject collaborative patterns. MQL~\citep{zhai2025mm-quant-gen} encodes multimodal item attributes (text, images) into a shared quantitative token space, enhancing cold-start and cross-domain performance. RL-based alignment~\citep{lu2024aligning-llm-rl} further improves controllability by optimizing instruction-following behavior with preference-based rewards, enabling conversational~\cite{friedman2023leveraginglargelanguagemodels,li2019deepconversationalrecommendations,chen2019knowledgebasedrecommenderdialog,Kemper_2024,li2023gpt4rec,seqreq-programming-pretrained} and constraint-aware recommendation.

\paragraph{Item ID Integration and Hybrid Representations.}  
To avoid verbose item descriptions, several works embed item IDs directly into the LLM’s vocabulary. CoVE~\citep{zhang2025cove} expands the token set with unique item tokens, enabling single-token recommendations and compressed embeddings. CLLM4Rec~\citep{zhu2024ccllm4rec} extends this with both user and item tokens, combining soft and hard prompts to integrate collaborative semantics.  
These ID-augmented models improve efficiency and accuracy but risk “knowledge entanglement”: naive merging of ID and language tokens can cause interference, harming both recommendation accuracy and language fluency.

\subsection{Generative and Hybrid Recommender Models}
Generative recommenders recast recommendation as a sequence generation task~\citep{yang2025gr-llm}, unifying retrieval and ranking in one model.  
HSTU~\citep{zhai2024hstu} employs a Transformer-based transducer, scaling up to 1.5T parameters and achieving large offline and online gains, while demonstrating NLP-like scaling laws for recommendation. TIGER~\citep{rajput2023tiger} compresses item vocabularies via multi-code vector quantization. OneRec~\citep{deng2025onerec} unifies retrieval and ranking in an encoder–decoder Transformer with sparse Mixture-of-Experts (MoE)~\cite{shazeer2017outrageously,fedus2022switch,ma2018modeling,moe-rec-2,mome,Zhang_2024moe,wang2024homehierarchymultigateexperts} for capacity scaling and adds Iterative Preference Optimization for alignment. These approaches offer novelty, explainability, and unified modeling, but require heavy compute and careful fine-tuning strategies to retain collaborative memory.

\paragraph{Beyond Accuracy.}  
Extensions like MTGR~\citep{han2025mtgr} integrate hand-crafted features into generative architectures, while others focus on fairness, calibration, and bias mitigation in LLM-based recommenders~\citep{yang2025gr-llm}. The generative format naturally supports novelty and explanation generation, which can combat popularity bias and improve transparency, but system design remains challenging.

\subsection{Multimodal MoE LLMs}

More recently, MoE has been integrated directly into multimodal large language models (MLLMs) and large vision–language models (LVLMs) \cite{bao2022vlmounifiedvlp,shen2023scalingvlmmoe,diao2025eve2,deng2025bagel,xiong2025llava,wang2025damo,xiong2026phycritic,ganjdaneshnot,Shirkavand_2025_CVPR}. MoE-LLaVA~\citep{lin2024moellava} introduces a sparse MoE backbone for LLaVA~\citep{liu2023llava}-style LVLMs and proposes a three-stage MoE-tuning strategy that first builds a strong dense LVLM and then converts its feed-forward blocks into experts. The resulting MoE-LLaVA model achieves performance comparable to or better than substantially larger dense LLaVA variants, while activating only a fraction of the parameters per token and reducing visual hallucinations~\cite{rawal2025argus}.

Uni-MoE~\citep{li2025unimoe} targets unified multimodal LLMs that support a broad set of modalities and tasks, applying MoE layers to scale capacity while maintaining a single generalist model. Addressing task interference in instruction-tuned MLLMs, MoME (Mixture of Multimodal Experts) ~\citep{mome} decomposes the architecture into a Mixture of Vision Experts (MoVE) and a Mixture of Language Experts (MoLE). MoVE aggregates features from multiple vision encoders via an adaptive deformable transformation and an instruction-conditioned router, while MoLE inserts sparsely gated adapter experts into LLM layers.

\section{Experiments}\label{sec:app-experiments}

\subsection{Baselines}\label{app:baselines}
We benchmark \textbf{\abbr} against representative methods spanning classic sequence modeling and recent LLM-based recommenders, with an emphasis on baselines that add recommendation capability to LLMs.

\textbf{Early sequential modeling.}
\emph{GRU4Rec}~\citep{hidasi2015gru4rec} pioneers GRU-based session modeling; \emph{SASRec}~\citep{kang2018sasrec} introduces unidirectional self-attention; \emph{BERT4Rec}~\citep{sun2019bert4rec} adopts bidirectional masked modeling for sequences.

\textbf{Transformer extensions and self-supervision.}
\emph{FDSA}~\citep{zhang2019fdsa} enriches feature dependencies within Transformers, and \emph{S3-Rec}~\citep{zhou2020s3} pretrains with sequence-aware self-supervision.

\textbf{Representation design, multimodality, and framework-style comparatives.}
\emph{VQ-Rec}~\citep{hou2023vq-req} learns discrete item codes via vector quantization; \emph{MissRec}~\citep{wang2023missrec} explores multimodal pretraining and transfer; \emph{TIGER}~\citep{rajput2023tiger} formulates autoregressive retrieval over semantic IDs. Framework baselines that unify text and recommendation include \emph{P5/P5-CID}~\citep{geng2022RLP,hua2023p5-cid} and its multimodal extension \emph{VIP5}~\citep{geng2023vip5}. \emph{E4SRec}~\citep{li2023e4srec} targets efficient sequential recommendation with a largely frozen LLM. \emph{ReAT}~\citep{cao2024gdm-aligning} aligns LLMs to recommendation through auxiliary, recommendation-specific tasks. For completeness on small Amazon benchmarks, we also report \emph{CoVE}~\citep{zhang2025cove}.

\textbf{Our reproduced and controlled variants.}
To isolate architectural effects under identical capacity, tokenizer, and training budget, we implement three LLM-based variants on the \emph{same backbone} as \abbr: (i) \emph{ID Transformer} (item tokens only); (ii) \emph{Item-ID LLM + text-derived bias} (ID embeddings augmented with text features); and (iii) \emph{Item-LLM} (vocabulary expansion with explicit item text but no MoE). We also reproduce strong non-LLM and hybrid sequential baselines, including \emph{SASRec}~\citep{kang2018sasrec} and \emph{HSTU}~\citep{zhai2024hstu}. Unless stated otherwise, all LLM-based baselines are matched to \abbr in active parameter count and trained with the same token budget, optimizer, sequence length, and schedules.

\subsection{Datasets}\label{sec:app-datasets}

\begin{table}[t]
\centering
\captionof{table}{Statistics of Amazon datasets used.}
\resizebox{0.5\linewidth}{!}{
\begin{tabular}{l c r}
\hline
Dataset & Total sequences & Num items \\
\hline
Games        &    42259          & 13839   \\ 
Instruments  &    17112          & 6250    \\
Arts         &    22171          & 9416    \\ 
Sports       &    35598          & 18357   \\
Beauty       &    22363          & 12101   \\
Toys         &    35598          & 11924   \\ 
\hline
Books(23)    &    776370         & 495063  \\
Beauty(23)   &    729576         & 207649  \\
Toys(23)     &    432264         & 162035  \\
\hline
\end{tabular}
}
\label{tab:amazon_datasets}
\end{table}
We use public Amazon Dataset: Games, Intruments and Arts~\citep{ni2019amzn} as well as Sports, Beauty and Toys~\cite{mcauley2015amzn}. See Table~\ref{tab:amazon_datasets} for dataset statistics. We also train and evaluate on our in-house industrial-scale dataset with millions of users and tens of thousands of items.

\subsection{Preprocessing}
We take the preprocessed version of Games, Arts, and sports from ~\citet{zhai2025mm-quant-gen}. We take small Sports, Beauty and Toys from ~\citet{zhang2025cove}. We download 2023 amazon variants from the official website~\cite{hou2024bridginglanguageitemsretrieval}. Following previous work~\cite{rajput2023tiger}, we first filter out unpopular users and items with less than five interactions. Then, we create user behavior sequences based on the chronological order.  We use chronological leave-last-k splitting per user: last 1 for test, the preceding 1 for validation, and the remainder for training. Item text comes from title and categories.
Maximum item history length is 50 items (most recent first). Maximum total token length (items + text) is 1024. We truncate text first, then items if necessary to satisfy the context size. We pad shorter sequences to 1024 with a special pad token; attention masks prevent loss on padded positions. We take the final unpadded position for evaluation.

\subsection{Optimization and Evaluation}
Optimizer is AdamW (betas $(0.9, 0.9999)$, eps $1\mathrm{e}{-8}$, weight decay $1\mathrm{e}{-2}$). We use linear warmup of 3000 iterations, then a cosine decay learning rate schedule. We tune learning rate with a grid search over $\{1e-3, 1e-5, 1e-5\}$ for \abbr and baselines. Training runs with \texttt{bfloat16} on NVIDIA A100-80GB.  Batch size is 128. We use standard next-token objectives that minimizes the KL divergence between the data distribution and the distribution of the LLM. We report NDCG@10/50, HR@10/50, and MRR. Metrics are computed over the full catalog. We train for 200 epochs on small amazon datasets and for 50 epochs on larger amazon datasets. For text benchmarks we use \texttt{lm-eval-harness}~\cite{eval-harness}. We constrain the output space to the unseen token items for retrieval quality.

\subsection{\abbr Details}
\begin{enumerate}
    \item Experts per FFN block: 2 (ID expert + Text expert).
    \item Routing: static token-type routing (ID tokens → ID expert; text tokens → Text expert).    
    \item Shared components: attention, LayerNorms, positional embeddings.
    \item Expert widths: Text expert width = 1. ID expert width = 1 for ablations. Tuned for main tables.
    \item Placement: all-layers become MoE for ablations. last-k with 4,8, 16 is tuned for main results.
    \item Freezing Policy: For Table 1 experiments (Text analysis) and ablations, LLM backbone is frozen. In other small-scale runs we select the best among: freeze-all, freeze-text-expert-only, and freeze-attention-only. In industrial dataset, we freeze everything and only train the item experts and item embeddings.
    \item Factorized Embedding: On amazon datasets, instead of a single embedding table $E \in \mathbb{R}^{N_{\text{items}} \times d}$, we first project to a lower dimensional space and then to the model dimension to reduce embedding parameters $E = W_l \times W_u$ where $W_l \in \mathbb{R}^{N_{\text{items}} \times d_{mid}}$ and $W_u \in \mathbb{R}^{d_{mid} \times d}$.
    \item For main results (not ablations and not Table~\ref{tab:text-feat-ablation}), we warm up the item expert with item-only sequences for 20\% of epochs, then gradually mix in text tokens with a linear schedule. Ablations with LLM-based models and Table 1 do not use this warm-up to ensure fairness.
    
\end{enumerate}

\subsection{Results}

\subsubsection{Proprietary Results}
Table~\ref{tab:proprietary_by_dataset} shows the results on our industrial dataset.

\begin{table}
\centering
\captionof{table}{Results on our industrial dataset.}
\resizebox{0.6\linewidth}{!}{
\begin{tabular}{l *{3}{c}}
\toprule
\textbf{Method} &
\multicolumn{3}{c}{\textbf{Industrial  $\Delta(\%)$}} \\
\cmidrule(lr){2-4}
& NDCG@10 & HR@10 & MRR \\
\midrule
SASRec~(baseline)  & \multicolumn{3}{c}{---} \\
HSTU          & +10.5\% & +2.7\%  & +13.2\% \\
ID Transformer   & +21.1\% & +8.9\%  & +23.1\% \\
Title-LLM     & -81.8\% & -87.6\%  & -98.4\% \\
Text-Attr LLM  & +25.4\% & +14.1\% & +25.9\%  \\
Item-LLM      & +23.5\% & +13.0\% & +24.3\% \\
\abbr         & +27.1\% & +16.6\% & +31.2\% \\
\bottomrule
\end{tabular}
}
\label{tab:proprietary_by_dataset}
\end{table}

\begin{figure}[t]
  \centering
  \includegraphics[width=\linewidth]{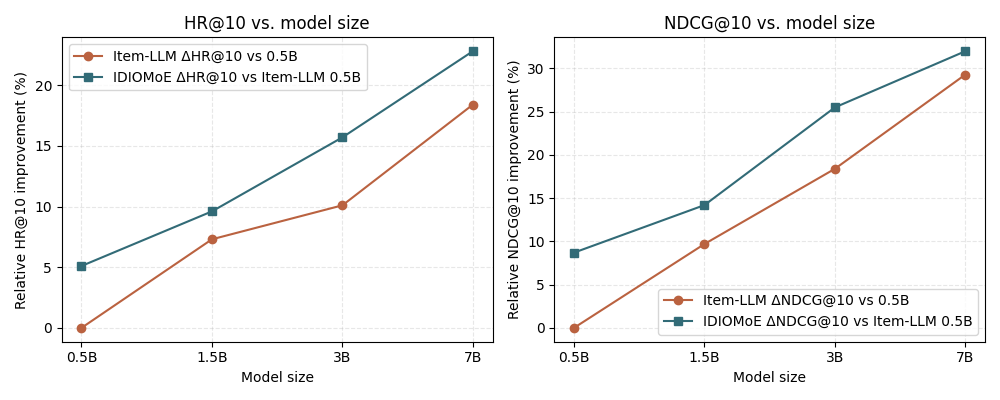}
  \caption{Relative performance on the industrial dataset with Qwen 2.5 backbones of different sizes. All values are reported as relative improvements (\%) over the 0.5B baseline Item-LLM.}
  \label{fig:scale-effect}
\end{figure}

We also conduct an additional experiement on our industrial dataset to study the effect of scaling the model using the Qwen 2.5~\cite{qwen2025qwen25technicalreport} family (0.5B, 1.5B, 3B, 7B). Figure~\ref{fig:scale-effect} shows the results. We see that recommendation quality improves with LLM size given enough training data, and the gains of \abbr over Item-LLM as the main baseline are persistent across all model scales considered.

\subsubsection{Semantic IDs}

\begin{table}[t]
\centering
\caption{Performance with Semantic IDs on three Amazon datasets.}
\resizebox{0.5\linewidth}{!}{
\begin{tabular}{l cc cc cc}
\toprule
& \multicolumn{2}{c}{Arts} & \multicolumn{2}{c}{Games} & \multicolumn{2}{c}{Instruments} \\
\cmidrule(lr){2-3}\cmidrule(lr){4-5}\cmidrule(lr){6-7}
Method   & HR@10 & NDCG@10 & HR@10 & NDCG@10 & HR@10 & NDCG@10 \\
\midrule
Item-LLM & 0.0946 & 0.0658 & 0.0823 & 0.0481 & 0.0826 & 0.0622 \\
IDIOMoE  & 0.1018 & 0.0730 & 0.0880 & 0.0492 & 0.0917 & 0.0686 \\
\bottomrule
\end{tabular}
}
\label{tab:semantic_ids}
\end{table}

\abbr is fully compatible with semantic ID schemes for handling new items. We conduct an experiments where we replace raw item IDs with semantic IDs from MQL4GRec~\citep{zhai2025mm-quant-gen}, showing that \abbr’s gains persist in this setting (Table~\ref{tab:semantic_ids}).

We also conducted a cold-start experiment following the steps described in the section 4.3 of TIGER~\citep{rajput2023tiger}, where we remove 5\% of test items from the training data and report the test performance overall and over the unseen set items. We set the ratio of unseen items to seen items in the top-k items $\epsilon = 0.1$. Table~\ref{tab:cold_start} shows the results, demonstrating that our method extends naturally to standard cold-start mechanisms. 

\begin{table}[t]
\centering
\caption{Cold-start evaluation following TIGER: 5\% of test items are removed from training. We report overall test metrics (All) and metrics restricted to unseen items (Unseen) for three datasets.}
\resizebox{\linewidth}{!}{
\begin{tabular}{l cccc cccc cccc}
\toprule
& \multicolumn{4}{c}{Arts} & \multicolumn{4}{c}{Games} & \multicolumn{4}{c}{Instruments} \\
\cmidrule(lr){2-5}\cmidrule(lr){6-9}\cmidrule(lr){10-13}

& \multicolumn{2}{c}{All} & \multicolumn{2}{c}{Unseen} & \multicolumn{2}{c}{All} & \multicolumn{2}{c}{Unseen} & \multicolumn{2}{c}{All} & \multicolumn{2}{c}{Unseen} \\
\cmidrule(lr){2-3}\cmidrule(lr){4-5}\cmidrule(lr){6-7}\cmidrule(lr){8-9}\cmidrule(lr){10-11}\cmidrule(lr){12-13}

Method       & HR@10 & NDCG@10& HR@10 & NDCG@10 & HR@10 & NDCG@10 & HR@10 & NDCG@10 & HR@10 & NDCG@10 & HR@10 & NDCG@10 \\
\midrule
Backbone  & 0.0808 & 0.0618 & 0.0569 & 0.0395 & 0.0849 & 0.0534 & 0.0478 & 0.0332 & 0.0642 & 0.0433 & 0.0394 & 0.0249 \\
IDIOMoE   & 0.0892 & 0.0643 & 0.0547 & 0.0416 & 0.0941 & 0.0572 & 0.0541 & 0.0422 & 0.0877 & 0.0579 & 0.0580 & 0.0313 \\
\bottomrule
\end{tabular}
}
\label{tab:cold_start}
\end{table}

These results demonstrate that our method extends naturally to standard cold-start mechanisms and is compatible with semantic-ID-based handling of new items.

\subsubsection{Attention analysis on text-only prompts}\label{sec:attn-text-only}

We analyze the internal attention behavior of our Item LLM on text-only inputs. We use the same tokenizer and pretrained backbone as the deployed model, run the model on a set of text prompts, and compute summary statistics per layer. We compare (i) \abbr (ii) a freshly loaded pretrained backbone.

For each transformer layer, we average heads, mask padding, and re-normalize per query. We report:
\begin{enumerate}
    \item previous-token attention, $A[i,i-1]$ averaged over valid positions
    \item attention to the first token, $A[:,0]$
    \item the distance profile, $A[i,i-d]$ as a function of offset $d$
    \item the entropy of the attention distribution over keys per query, averaged over queries.
\end{enumerate}

We also aggregate distance profiles over early/mid/late layer blocks for clarity.

\begin{figure}[t]
  \centering
  \includegraphics[width=0.48\linewidth]{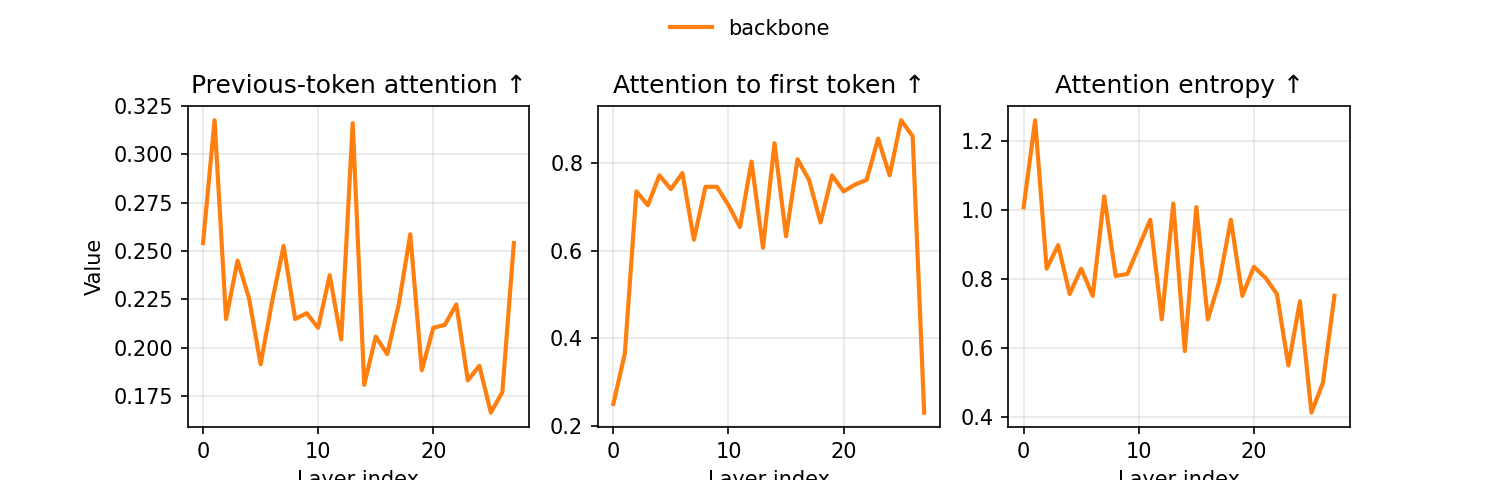}
  \includegraphics[width=0.48\linewidth]{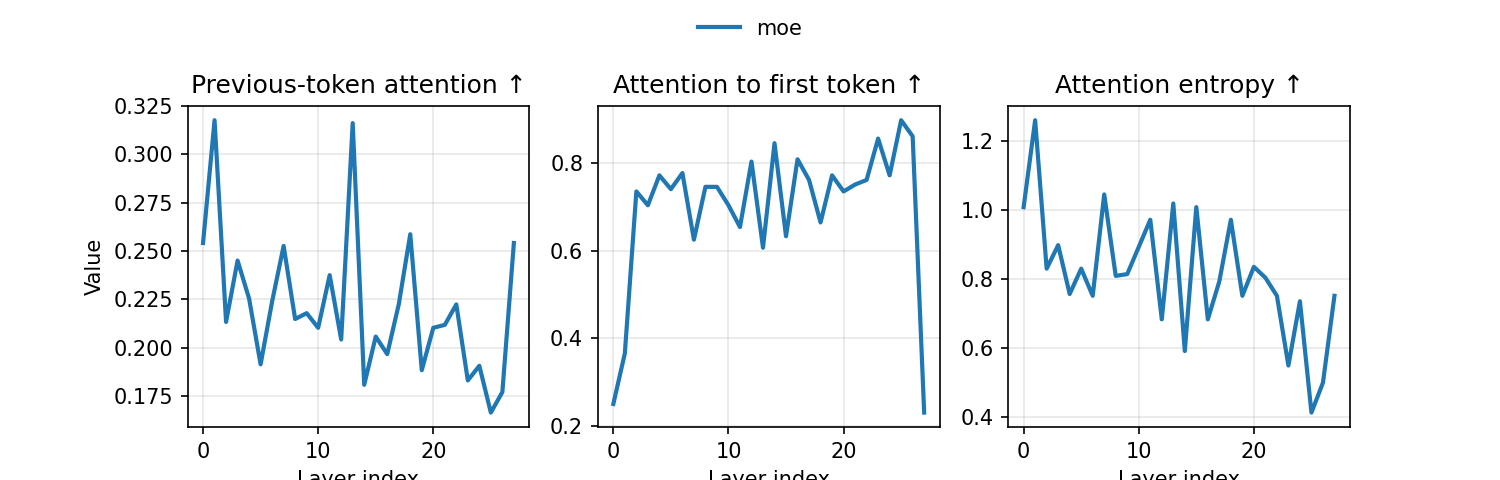}
  \caption{Layer-wise attention metrics on text-only inputs. Left: previous-token attention. Middle: attention to the first token. Right: attention entropy. MoE (blue) and backbone (orange) overlap across layers, indicating preserved attention geometry.}
  \label{fig:attn-layer-trends}
\end{figure}

\begin{figure}[t]
  \centering
  \includegraphics[width=0.48\linewidth]{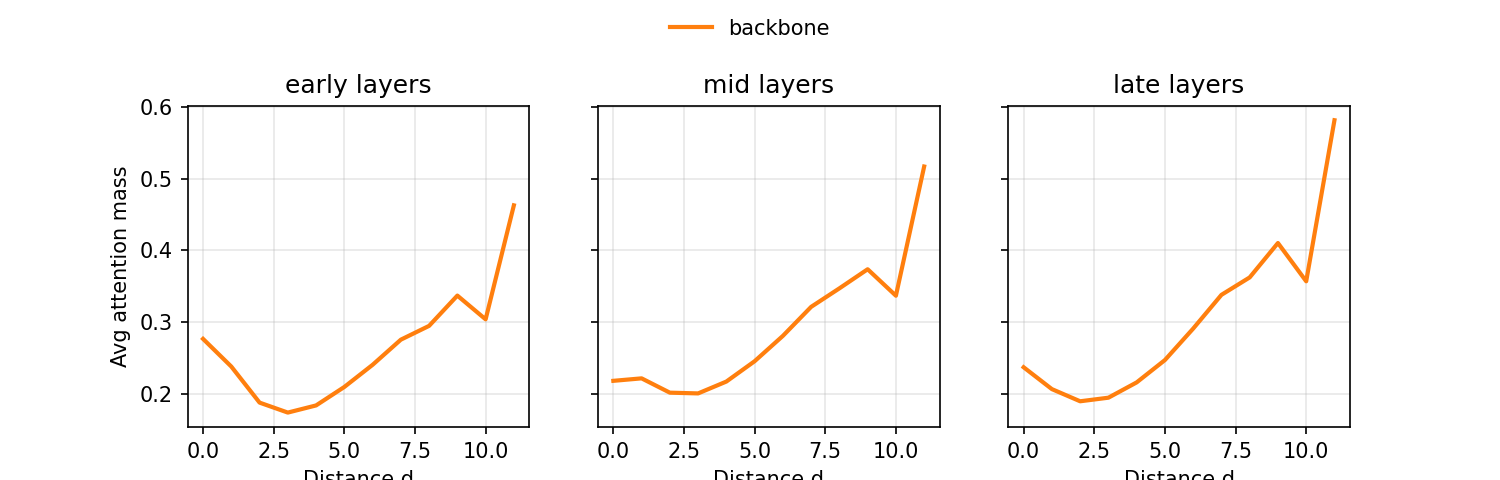}
  \includegraphics[width=0.48\linewidth]{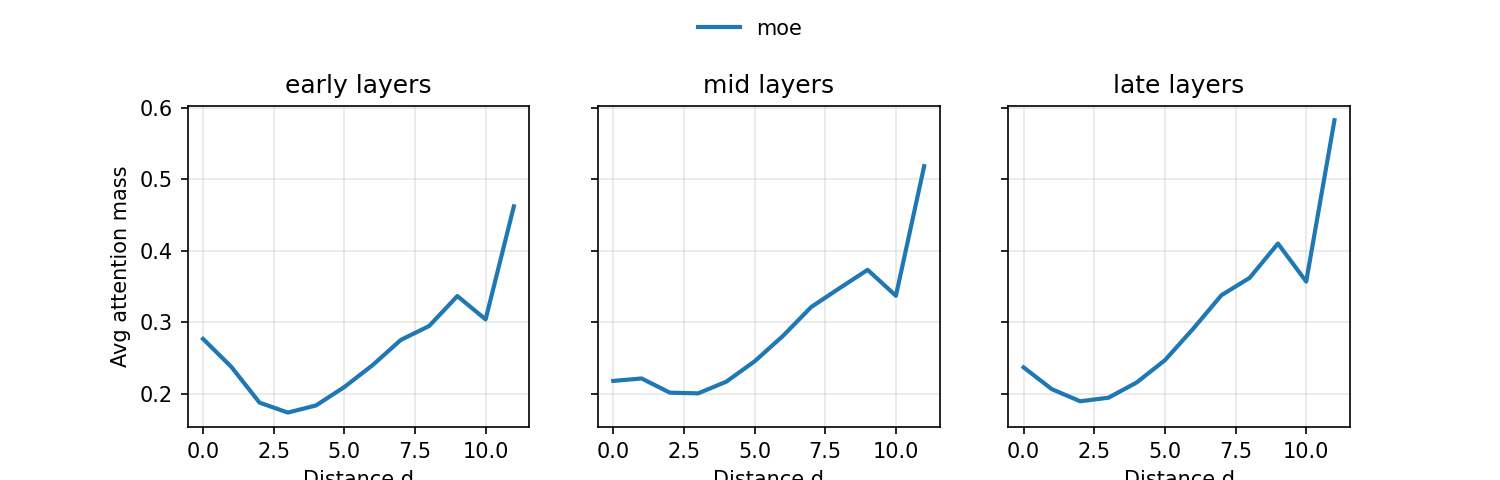}
  \caption{Distance profiles aggregated over early, mid, and late layers. MoE and backbone curves are nearly identical, reflecting similar allocation of attention mass across short-, medium-, and long-range dependencies.}
  \label{fig:attn-distance-profiles}
\end{figure}

Figures~\ref{fig:attn-layer-trends} and~\ref{fig:attn-distance-profiles} show that the MoE model and the pretrained backbone exhibit \emph{near-identical} attention patterns on text-only inputs across all layers. Layer-wise previous-token bias, first-token emphasis, and attention entropy overlap almost perfectly, and early/mid/late distance profiles coincide within visual resolution.

This alignment is expected in our setting for two reasons:
\begin{enumerate}
    \item The MoE architecture modifies the feed-forward pathways, while the backbone self-attention blocks remain architecturally unchanged
    \item The text-only inputs do not activate item-specific experts, so the effective computation path closely matches the backbone.
\end{enumerate}
Consequently, attention \emph{structure} (diagonal strength, range of contextual aggregation) remains stable, even though token-level representations downstream of attention can still differ due to MoE expert routing within the MLPs.
Under text-only prompts, our fine-tuned Item LLM preserves the backbone's attention geometry. This suggests that improvements from MoE primarily arise in representation and computation within expert MLPs rather than from altering attention allocation. 

\subsubsection{Efficiency Results}\label{app:efficiency-results}

\rebuttal{
Our model is evaluated with standard batched inference. It is not restricted to processing a single query at a time. Just like a conventional LLM, \abbr supports multi-query batches with appropriate padding and attention masking, and all of our reported results use evaluation batch sizes larger than 1. 
Table~\ref{tab:efficiency_itemratio02_b8} reports latency and throughput values for both batched training and inference on three sequence lengths, showing that \abbr achieves comparable performance values to the underlying backbone model at various sequence lengths. The MoE modification only changes the FFN sublayers (attention remains shared), so the per-token compute remains similar, and there is no additional online adaptation step at serving time beyond a single forward pass. From Table~\ref{tab:efficiency_itemratio02_b8} two trends stand out:
}
\begin{enumerate}
    \item Overhead shrinks with sequence length. At short contexts (256 tokens), MoE adds modest training overhead (+6.5\% latency, -6.1\% tokens/s) and a larger inference overhead (+18.4\% latency). As context grows, routing/pack–scatter costs amortize: at 512 tokens the inference overhead drops to +12.5\%, and at 1024 tokens it is only +3.8\% with no memory increase. Training overhead is similarly small at long sequences ($\leq 0.7\%$ tokens/s at 1024).
    \item Memory is neutral. Peak GPU memory is within $\pm0.5$G of the dense baseline across all settings, and identical at 1024 tokens for both training (29.4G) and inference (4.67G), consistent with activating one expert per token.    
\end{enumerate}

\abbr achieves near-parity efficiency at long contexts ($\leq 4\%$ overhead at 1024) and acceptable overheads at short contexts ($\approx18\%$ at 256), while keeping memory effectively unchanged. In Section~\ref{sec:experiments}, we show these costs buy consistent quality gains placing \abbr on a favorable quality–latency Pareto frontier.

\begin{table}[t]
\centering
\caption{Efficiency at batch size $8$ for three sequence lengths with item ratio of $0.2$. $\Delta$ is MoE relative to the dense baseline. Latency is end-to-end per query; throughput is steady-state.}
\resizebox{\linewidth}{!}{
\begin{tabular}{lcccccccccccccc}
\toprule
\multirow{2}{*}{Seq} & \multirow{2}{*}{Phase} &
\multicolumn{3}{c}{Latency (ms) $\downarrow$} & \multicolumn{3}{c}{Examples/s $\uparrow$} & \multicolumn{3}{c}{Tokens/s $\uparrow$} & \multicolumn{3}{c}{Peak Mem (G) $\downarrow$} \\
\cmidrule(lr){3-5}\cmidrule(lr){6-8}\cmidrule(lr){9-11}\cmidrule(lr){12-14}
 & & Base & MoE & $\Delta$ & Base & MoE & $\Delta$ & Base & MoE & $\Delta$ & Base & MoE & $\Delta$ \\
\midrule
\multirow{2}{*}{256}
& Train & 117.86 & 125.53 & +6.5\% & 67.88 & 63.73 & $-6.1\%$ & 17377.08 & 16314.61 & $-6.1\%$ & 10.45 & 10.51 & +0.6\% \\
& Infer & 36.13  & 42.78  & +18.4\% & 221.44 & 186.99 & $-15.6\%$ & 56689.83 & 47870.29 & $-15.6\%$ & 2.81 & 2.82 & +0.4\% \\
\midrule
\multirow{2}{*}{512}
& Train & 180.59 & 186.58 & +3.3\% & 44.30 & 42.88 & $-3.2\%$ & 22681.76 & 23196.24 & {+2.3\%} & 16.72 & 16.80 & +0.5\% \\
& Infer & 49.16  & 55.33  & +12.5\% & 162.72 & 144.59 & $-11.2\%$ & 83314.50 & 74028.48 & $-11.2\%$ & 3.43 & 3.43 & {0.0\%} \\
\midrule
\multirow{2}{*}{1024}
& Train & 323.48 & 323.98 & +0.2\% & 24.73 & 24.69 & $-0.2\%$ & 25324.61 & 25146.42 & $-0.7\%$ & 29.40 & 29.40 & {0.0\%} \\
& Infer & 92.45  & 95.92  & +3.8\% & 86.53  & 83.40  & $-3.6\%$ & 88607.00 & 85400.26 & $-3.6\%$ & 4.67 & 4.67 & {0.0\%} \\
\bottomrule
\end{tabular}
}
\label{tab:efficiency_itemratio02_b8}
\end{table}

\end{document}